\pdfoutput=1
\documentclass{article}

\usepackage{arxiv}

\usepackage{microtype}
\usepackage{graphicx}
\usepackage{subfigure}
\usepackage{booktabs} 

\usepackage{amsmath,amsfonts,bm}

\def\eqref#1{equation~\ref{#1}}

\def\1{\bm{1}}

\DeclareMathAlphabet{\mathsfit}{\encodingdefault}{\sfdefault}{m}{sl}
\SetMathAlphabet{\mathsfit}{bold}{\encodingdefault}{\sfdefault}{bx}{n}

\usepackage{authblk}
\usepackage{hyperref}
\usepackage{multirow}
\usepackage{natbib}
\usepackage{url}
\usepackage{booktabs}
\usepackage{tikz}
\usepackage{wrapfig}
\usepackage{subcaption} 
\usetikzlibrary{matrix,decorations.pathreplacing,arrows.meta,positioning, calc, shapes.geometric, backgrounds}
\definecolor{lavender}{rgb}{0.96, 0.73, 1.0}
\definecolor{brickred}{rgb}{0.8, 0.25, 0.33}
\newcommand{\std}[1]{\textcolor{gray}{\(\pm\) \footnotesize{#1}}}
\definecolor{mygreen}{RGB}{0, 153, 0}

\usepackage{hyperref}

\usepackage{amsmath}
\usepackage{amssymb}
\usepackage{mathtools}
\usepackage{amsthm}

\usepackage[capitalize,noabbrev]{cleveref}

\theoremstyle{plain}

\theoremstyle{definition}

\theoremstyle{remark}

\usepackage[textsize=tiny]{todonotes}

\title{Mambular: A Sequential Model for Tabular Deep Learning}

\author[1]{\textbf{\small Anton Frederik Thielmann\thanks{Correspondence to \texttt{anton.thielmann@basf.com}}}}
\author[1]{\textbf{\small Manish Kumar}}
\author[1]{\textbf{\small Christoph Weisser}}
\author[2]{\textbf{\small Arik Reuter}} 
\author[3]{\textbf{\small Benjamin~Säfken}} 
\author[4]{\textbf{\small Soheila Samiee}}

\affil[1]{\small BASF, Germany}
\affil[2]{\small LMU Munich, Germany}
\affil[3]{\small Clausthal University of Technology, Germany}
\affil[4]{\small BASF, Canada}

\begin{document}
\maketitle

\begin{abstract}
The analysis of tabular data has traditionally been dominated by gradient-boosted decision trees (GBDTs), known for their proficiency with mixed categorical and numerical features. However, recent deep learning innovations are challenging this dominance. This paper investigates the use of autoregressive state-space models for tabular data and compares their performance against established benchmark models. 
Additionally, we explore various adaptations of these models, including different pooling strategies, feature interaction mechanisms, and bi-directional processing techniques to understand their effectiveness for tabular data. Our findings indicate that interpreting features as a sequence and processing them and their interactions through structured state-space layers can lead to significant performance improvement. 
This research underscores the versatility of autoregressive models in tabular data analysis, positioning them as a promising alternative that could substantially enhance deep learning capabilities in this traditionally challenging area. 
The source code is available at \url{https://github.com/basf/mamba-tabular}.
\end{abstract}

\section{Introduction}

Gradient-boosted decision trees (GBDTs) have long been the dominant approach for analyzing tabular data, due to their ability to handle the typical mix of categorical and numerical features found in such datasets \citep{grinsztajn2022tree}.  In contrast, deep learning models have historically faced challenges with tabular data, often struggling to outperform GBDTs. The complexity and diversity of tabular data, including issues like missing values, varied feature types, and the need for extensive preprocessing, have made it difficult for deep learning to match the performance of GBDTs \citep{borisov2022deep}. However, recent advancements in deep learning are gradually challenging this paradigm by introducing innovative architectures that leverage advanced mechanisms to capture complex feature dependencies, promising significant improvements \citep{popov2019neural,hollmann2022tabpfn, gorishniy2021revisiting}.

One of the most effective advancements in tabular deep learning is the application of attention mechanisms in models like TabTransformer \citep{huang2020tabtransformer}, FT-Transformer \citep{gorishniy2021revisiting} and many more \citep{wang2022transtab, thielmanninterpretable, arik2021tabnet}. These models leverage the attention mechanism to capture dependencies between features, offering a significant improvement over traditional approaches. FT-Transformers, in particular, have demonstrated robust performance across various tabular datasets, often surpassing the accuracy of GBDTs \citep{mcelfresh2024neural}. 
Additionally, more traditional models like Multi-Layer Perceptrons (MLPs) and ResNets have demonstrated improvements when well-designed and when the data undergoes thorough preprocessing \citep{gorishniy2021revisiting, gorishniy2022embeddings}. These models have benefited especially from innovations in advanced preprocessing methods that make them more competitive.

More recently, the Mamba architecture \citep{gu2023mamba} has shown promising results in textual problems. Tasks previously dominated by Transformer architectures, such as DNA modeling and language modeling, have seen improvements with the application of Mamba models \citep{gu2023mamba, schiff2024caduceus, zhao2024cobra}.
Several adaptations have demonstrated its versatility, such as Vision Mamba for image classification \citep{xu2024survey}, video analysis \citep{yang2024vivim, yue2024medmamba} and point cloud analysis \citep{zhang2024point, liu2024point}. Furthermore, the architecture has been adapted for time series problems, with notable successes reported by \citet{patro2024simba}, \citet{wang2024mamba} and \citet{ahamed2024timemachine}. 
Mamba has also been integrated into graph learning \citep{behrouz2024graph} and imitation learning \citep{correia2024hierarchical}. 
Further advancements have improved the language model, for example, by incorporating attention \citep{lieber2024jamba}, Mixture of Experts \citep{pioro2024moe} or bi-directional sequence processing \citep{liang2024bi}.

These advancements underscore Mamba's broad applicability, making it a powerful and flexible architecture for diverse tasks and data types. Similarly to the transformer architecture, the question arises whether the Mamba architecture can also be leveraged for tabular problems, and this study is focused on addressing this question. 

The contributions of the paper can be summarized as follows:
\begin{enumerate}
\renewcommand{\labelenumi}{\Roman{enumi}.}
    \item We introduce a tabular adaptation of Mamba, subsequently called \textit{Mambular}, demonstrating the potential of sequential models in addressing tabular problems.
    \item  We conduct extensive benchmarking of Mambular against several competitive neural and tree-based methods, illustrating that a standard Mambular model performs on par with or better than tree-based models across a wide range of datasets.
    \item We examine the impact of bi-directional processing and feature interaction layers on Mambular's performance, and compare several pooling methods.
    \item Finally, we carry out an in-depth analysis of the models autoregressive structure, investigating the implications of feature orderings in a sequential tabular model.
\end{enumerate}

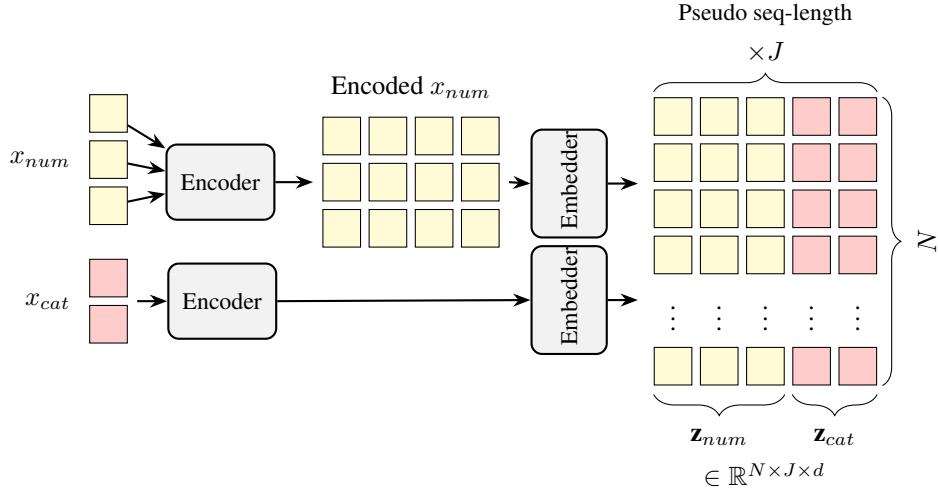
\begin{figure*}[ht]
\centering
\begin{tikzpicture}[
    node distance=0.2cm and 0.5cm, 
    feature/.style={draw, thick, rectangle, fill=yellow!20, minimum width=1.5cm, minimum height=0.5cm},
    arrow/.style={-Stealth, thick},
    block/.style={draw, rounded corners, thick, fill=gray!10, text width=1.2cm, align=center, minimum height=1cm},
    matrixstyle/.style={matrix of nodes, nodes={draw, minimum size=5mm, fill=yellow!20}, column sep=1mm, row sep=1mm, nodes in empty cells},
    transformer/.style={draw, rounded corners, thick, fill=gray!10, text width=1.2cm, align=center, minimum height=1cm, rotate=90},
]

\matrix (numFeatures) [matrixstyle, label=left:$x_{num}$] {
    \\
    \\
    \\
};

\matrix (catFeatures) [matrixstyle, below=of numFeatures, label=left:$x_{cat}$, nodes={fill=red!20}] {
    \\
    \\
};

\node[block, right=of numFeatures-2-1, yshift=-0.3cm] (basis) {\small{Encoder}};

\matrix (numFeatureMatrix) [matrixstyle, right=of basis, label=above: Encoded $x_{num}$] {
    & & &\\
    & & &\\
    & & &\\
};

\node[transformer, right=of numFeatureMatrix, xshift=-0.75cm, yshift=-0.3cm](numtok){\small Embedder};

\node[block, right=of catFeatures, xshift=-0.1cm, yshift=0cm] (tokenizer) {\small Encoder};
\node[transformer, right=of numFeatureMatrix, xshift=-2.3cm, yshift=-0.3cm] (cattok) {\small Embedder};

\matrix (allFeatures) [matrixstyle, right=of numtok, yshift=-1.5cm, xshift=0.5cm, nodes={fill=none}] {
    |[fill=yellow!20]| & |[fill=yellow!20]|& |[fill=yellow!20]|& |[fill=red!20]|&|[fill=red!20]|\\
    |[fill=yellow!20]| & |[fill=yellow!20]|& |[fill=yellow!20]|& |[fill=red!20]|&|[fill=red!20]|\\
    |[fill=yellow!20]| & |[fill=yellow!20]|& |[fill=yellow!20]|& |[fill=red!20]|&|[fill=red!20]|\\
    |[fill=yellow!20]| & |[fill=yellow!20]|& |[fill=yellow!20]|& |[fill=red!20]|&|[fill=red!20]|\\
    \node[draw=none] {$\vdots$}; & \node[draw=none] {$\vdots$}; & \node[draw=none] {$\vdots$}; & \node[draw=none] {$\vdots$}; & \node[draw=none] {$\vdots$}; \\
    |[fill=yellow!20]| & |[fill=yellow!20]|& |[fill=yellow!20]|& |[fill=red!20]|&|[fill=red!20]|\\
};

\draw [decorate, decoration={brace, amplitude=7pt, raise=2pt}]
    (allFeatures-1-5.north east) -- (allFeatures-6-5.south east) 
    node [midway, right=12pt, anchor=north, rotate=90] {$N$};

\draw [decorate, decoration={brace, amplitude=7pt, raise=2pt}] 
    (allFeatures-1-1.north west) -- (allFeatures-1-5.north east) 
    node [midway, yshift=0.6cm] {$\times J$}
    node [midway, yshift=1.1cm] {\footnotesize{Pseudo seq-length}};

\draw [decorate, decoration={brace, amplitude=8pt, raise=18pt, mirror}] 
    (allFeatures-6-1.north west) -- (allFeatures-6-3.north east) 
    node [midway, yshift=-1.2cm] {$\textbf{z}_{num}$}
    node [midway, yshift=-1.7cm, xshift=0.6cm] {$\in \mathbb{R}^{N\times J \times d}$};

\draw [decorate, decoration={brace, amplitude=8pt, raise=18pt, mirror}] 
    (allFeatures-6-4.north west) -- (allFeatures-6-5.north east) 
    node [midway, yshift=-1.2cm] {$\textbf{z}_{cat}$};

\foreach \i in {1,2,3} {
    \draw[arrow] (numFeatures-\i-1) -- (basis);
}

\draw[arrow] (basis) -- (numFeatureMatrix);

\draw[arrow] (numFeatureMatrix.east) -- (numtok.north);
\draw[arrow] (catFeatures.east) -- (tokenizer.west);
\draw[arrow] (tokenizer.east) -- (cattok.north);

\draw[arrow] (numtok.south) -- ([xshift=0.4cm]numtok.south);

\draw[arrow] (cattok.south) -- ([xshift=0.4cm]cattok.south);

\end{tikzpicture}
\caption{Generation of the input matrix that are fed through the Mamba blocks. The categorical features are tokenized and embedded similar to classical embeddings for language models. The numerical features are encoded and embedded via a simple linear layer. The final input matrix of the Mamba blocks are the concatenated embeddings  $\mathbf{z} \in \mathbb{R}^{N \times J \times d}$ with embedding dimension $d$.}
\label{fig:feature_processing}
\end{figure*}

\section{Methodology}
For a tabular problem, let $\mathcal{D} = \{ (\bm{x}^{(i)}, y^{(i)})\}_{i=1}^n$ be the training dataset of size $n$ and let $y$ denote the target variable that can be arbitrarily distributed. Each input $\bm{x} = (x_1, x_2, \dots, x_J)$ contains $J$ features (variables).
Categorical and numerical features are distinguished such that $\bm{x} \equiv (\bm{x}_{cat}, \bm{x}_{num})$, with the complete feature vector denoted as $\bm{x}$.
Further, let $x_{j(cat)}^{(i)}$ denote the $j$-th categorical feature of the $i$-th observation, and hence $x_{j(num)}^{(i)}$ denote the $j$-th numerical feature of the $i$-th observation.

Following standard tabular transformer architectures, the categorical features are first encoded and embedded. In contrast to classical language models, each categorical feature has its own, distinct vocabulary to avoid problems with binary or integer encoded variables. Including \textsc{$<$unk$>$} tokens additionally allows to easily deal with unknown or missing categorical values during training or inference.

Numerical features are mapped to the embedding space via a simple linear layer. However, since a single linear layer does not add information beyond a linear transformation, Periodic Linear Encodings, as introduced by \citet{gorishniy2022embeddings} are used for all numerical features. Thus, each numerical feature is encoded before being passed through the linear layer for rescaling. Simple decision trees are used for detecting the bin boundaries, $b_t$, and depending on the task, either classification or regression is employed for the target-dependent encoding function $h_j(\bm{x}_{j(num)}, y)$. Let $b_t$ denote the decision boundaries from the decision trees. The encoding function is given in Eq. \ref{eq:ple}.

\begin{center}
    \textbf{PLE}
\end{center}
\begin{equation}\label{eq:ple}
z_{j(\text{num})}^{t} = \begin{cases} 
0 & \text{if } x < b_{t-1}, \\
1 & \text{if } x \geq b_t, \\
\frac{x - b_{t-1}}{b_{t-2} - b_{t-1}} & \text{else.}
\end{cases}
\end{equation}

The feature encoding and embedding generation is demonstrated in Figure \ref{fig:feature_processing}. The created embeddings, following classical statistical literature \citep{hastie2009elements, Kneib} are denoted as $\mathbf{Z}$ and not $\mathbf{X}$ to clarify the difference between the embeddings and the raw features.

Subsequently, the embeddings are passed jointly through a stack of Mamba layers. These include one-dimensional convolutional layers to account for invariance of feature ordering in the pseudo-sequence as well as a state-space (SSM) model \citep{gu2021efficiently, hamilton1994state} . The feature matrix before being passed through the SSM model has a shape of \textsc{(Batch Size) $\times$ J $\times$ (Embedding Dimension)}, later referenced as $N \times J \times d$. Importantly, the sequence length in a tabular context refers to the number of variables, and hence the second dimension, $J$, corresponds to the number of features rather than to the length of, e.g., a document.

The convolution operation along the sequence length \( J \) and with Kernel $K$ is expressed as:
\begin{equation*}
    \begin{aligned}
        \mathbf{Z}_{\text{conv}}^{(n,d)}(j) &= \sum_{m=0}^{K-1} \mathbf{Z}^{(n,d)}[j+m] \cdot \mathbf{k}^{(d)}(m), \\
        \forall n &\in \{1, \dots, N\}, \\
        \forall d &\in \{1, \dots, d\}, \\
        \forall j &\in \{1, \dots, J-K+1\}.
    \end{aligned}
\end{equation*}

where \( \mathbf{Z}_{\text{conv}}^{(n,d)}(j) \) is the \( j \)-th element of the convolved sequence for batch \( n \) and feature channel \( d \). $\mathbf{Z}^{(n,d)}[j+m] $ is the \( [j+m] \)-th element of the input sequence \( \mathbf{Z} \) for batch \( n \) and feature channel \( d \), and \( K \) describes the kernel size. Summing over the elements of the kernel, indexed by \( m \), accounts for the variable position in the pseudo-sequence. Thus, setting the kernel size equivalent to the number of variables would make the sequence invariant positional permutations.
The resulting output tensor retains the same shape as the input, since padding is set to the kernel size -1.

After the convolution, given the matrices:
\[
\mathbf{A} \in \mathbb{R}^{1 \times 1 \times d \times \delta}, \quad
\mathbf{B} \in \mathbb{R}^{N \times J \times 1 \times \delta}, \quad
\Delta \in \mathbb{R}^{N \times J \times d \times 1}\]
\[ \bar{\mathbf{z}} \in \mathbb{R}^{N \times J \times d \times 1}
,\]
where $\delta$ denotes a inner dimension, similar to the feed forward dimension in Transformer architectures and $\bar{\mathbf{z}}$ has the same entries as $\mathbf{z}$, but one additional axis, the formula for updating the hidden state \(\mathbf{h}_j \in \mathbb{R}^{N \times d\times \delta}\) is:
\begin{equation}\label{eq:ssm}
\begin{aligned}
\mathbf{h}_j = \exp\left(\Delta \odot_3 \mathbf{A} \right)_{:, j, :, :}
\odot_{1,2,3} \mathbf{h}_{j-1} + \\ \left (\left( \Delta \odot_{1,2} \mathbf{B} \right) \odot_{1,2,3} \bar{\mathbf{z}}\right )_{:, j, :, :}.
\end{aligned}
\end{equation}
The symbol $\odot_{d}$ denotes an outer product where the multiplication is done for the $d$-th axis and parallelized wherever a singleton axis length meets an axis of length one\footnote{This corresponds to using the ordinary multiplication operator "*" in PyTorch and relying on the default broadcasting}. The exponential function is applied element-wise.
The state transition matrix $\mathbf{A}$ governs the transformation of the hidden state from the previous time step to the current one, capturing how the hidden states evolve independently of the input features. The input-feature matrix  $\mathbf{B}$ maps the input features to the hidden state space, determining how each feature influences the hidden state at each step. The gating matrix  $\mathbf{\Delta}$ acts as a gating mechanism, modulating the contributions of the state transition and input-feature matrices, and allowing the model to control the extent to which the previous state and the current input affect the current hidden state. 
\begin{figure}[ht]
\centering
\begin{tikzpicture}
    \tikzset{
        box/.style={rectangle, draw, minimum width=0.9cm, minimum height=1.1cm, rounded corners, thick},
        arrow/.style={-Stealth, thick},
        dashedarrow/.style={-Stealth, thick, dashed},
        input/.style={rectangle, draw=none, fill=none, text=red, minimum width=1.3cm},
        output/.style={rectangle, draw=none, fill=none, text=purple, minimum width=1.3cm}
    }

    \node[box, shading=axis, left color=yellow!20, right color=red!30] (x) {$z_j$};
    \node[box, right=of x, fill=gray!10] (B) {$B$};
    \node[box, right=of B, fill=blue!20] (h) {$h_j$};
    \node[box, below=of h, fill=blue!20, node distance=2cm] (A) {$A h_{j-1}$};
    \node[box, right=of h, fill=gray!10] (C) {$C$};
    \node[box, fill=green!20, right=of C] (y) {$\tilde{x}_j$};
    \node[box, above=of h, fill=gray!10, node distance=2cm] (D) {$\mathbf{\alpha}$};

    \draw[arrow] (x) -- (B);
    \draw[arrow] (B) -- (h);
    \draw[arrow] (h) -- (C);
    \draw[arrow] (C) -- (y);
    
    \draw[arrow] (h.west) to[out=225, in=135] (A.west);
    \draw[arrow] (A.east) to[out=45, in=-45] (h.east);

    \draw[arrow] (x.north) to[out=45, in=180] (D.west);
    \draw[arrow] (D.east) to[out=0, in=135] (y.north);

    \draw[dashedarrow] (x.south) to[out=-45, in=-135] (B.south);
    \draw[dashedarrow] (x.north) to[out=15, in=155] (C.north);

\end{tikzpicture}
\caption{SSM updating step with recursive update of $h$: The hidden state is iteratively updated by going through the sequence (features) similar to a recurrent neural network. The final representation is generated as described in Equations 3-4.}
\label{fig:mamba_arch}
\end{figure}
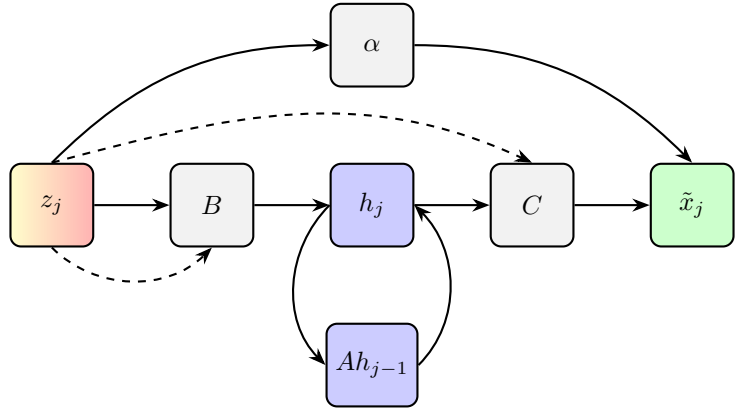

In contrast to FT-Transformer \citep{gorishniy2021revisiting} and TabTransformer \citep{huang2020tabtransformer} Mambular truly iterates through all variables as if they are a sequence; hence, feature interactions are detected sequentially. The effect of feature position in a sequence, and the impact of the convolution kernel size is analyzed with respect to performance in section \ref{seq:ablation}. 
Furthermore, it should be noted that in contrast to TabPFN \citep{hollmann2022tabpfn}, Mambular does not transpose dimensions and iterates over observations. Hence, training on large datasets is possible and it can scale well to any training data size, just as Mamba \citep{gu2023mamba} does.

After stacking and further processing, the final representation, $\tilde{\textbf{x}} \in \mathbb{R}^{N \times J \times d}$ is retrieved. In truly sequential data, these are the contextualized embeddings of the input tokens, for tabular problems $\tilde{\textbf{x}}$ represents a contextualized, or feature interaction accounting variable representation, in the embedding space.
The hidden states are stacked along the sequence dimension to form:
\[
\mathbf{H} = [\mathbf{h}_0, \mathbf{h}_1, \ldots, \mathbf{h}_{T-1}] \in \mathbb{R}^{N \times J \times d \times \delta}.
\]

The final output representation \(\tilde{\mathbf{x}}\) is then computed by performing matrix multiplication of the stacked hidden states with matrix \(\mathbf{C}\ \in \mathbb{R}^{N \times J \times 1 \times \delta}\) where the multiplication and summation is done over the last axis, and adding the vector \(\mathbf{\alpha}\ \in \mathbb{R}^{1 \times 1 \times d} \) scaled by the input \(\mathbf{z}\):
\begin{equation}
\tilde{\mathbf{x}} = \left(  \mathbf{H} \cdot_4 \mathbf{C} \right) + \left( \mathbf{\alpha} \odot_3 \mathbf{z} \right).
\end{equation}

More explicitly, this can be written as:
\begin{equation*}
\tilde{x}_{i,j,k} = \sum_{\delta} \mathbf{H}_{i,j,k,\delta} \mathbf{C}_{i,j,1,\delta} + \mathbf{\alpha}_{1,1,k} \mathbf{z}_{i,j,k}.
\end{equation*}

where \(\mathbf{C}\) and \(\mathbf{\alpha}\) are learnable parameters.
For final processing, \(\tilde{\mathbf{x}}\) is element-wise multiplied with \(\mathbf{z}'\), and the result is passed through a final linear layer:
\begin{equation}\label{eq:final}
\tilde{\mathbf{x}}_{\text{final}} = (\tilde{\mathbf{x}} \odot_{1,2,3} \mathbf{z'}) \mathbf{W}_{\text{final}} + \mathbf{b}_{\text{final}}.
\end{equation}

Before passing \(\tilde{\mathbf{x}}_{\text{final}}\) to the final task-specific model head, pooling is applied along the feature axis. Various strategies are analyzed in Section~\ref{sec:ablation}, with average pooling used as the default.

\begin{figure}[ht]
\centering

\begin{tikzpicture}[
  node distance=0.4cm and 0cm,
  every node/.style={draw, very thick, rounded corners, minimum width=2cm, minimum height=0.9cm, align=center, font=\small},
  arrow/.style={->, thick},
  background rectangle/.style={fill=gray!20, rounded corners}
]

\node (input) [draw, fill=yellow!20] {Input Sequence $\mathbf{x}$};
\node (encoding) [ellipse, draw,minimum width=1.5cm, minimum height=1cm,  right=of input,  yshift=0.9cm, label=center:{Encoding}] {};
\node (embedding) [above=of input, yshift=0.25cm, draw, fill=yellow!20] {Embedding Layer};

\node (conv) [above=of embedding, draw, fill=red!20] {1D Convolution \\ Linear Layer};
\node (x_part) [above left=0.3cm and -0.3cm of conv, draw, fill=red!20, minimum height=0.5cm] {$\mathbf{z}$};
\node (z_part) [above right=0.3cm and -0.3cm of conv, draw, fill=red!20, minimum height=0.5cm] {$\mathbf{z'}$};

\node (A-matt) [above= of z_part, draw, fill=gray!20] {$\mathbf{A}$ \\ $\mathbf{B}$ \\ $\mathbf{\Delta}$};
\node (D-matt) [above= of A-matt, draw, fill=gray!20] {$\mathbf{\alpha}$ \\ $\mathbf{C}$};

\node (hidden_update) [above=of x_part, draw, fill=blue!20] {SSM};
\node (ssm) [ellipse, draw,minimum width=1.9cm, minimum height=1cm,  fill=blue!20, left=of hidden_update,  xshift=-0.7cm, yshift=1.0cm, rotate=90, label=center:{Update \\ Eq. 2}] {};
\node (elementwise_mult) [above=of hidden_update, draw, fill=green!20] {$\tilde{\mathbf{x}} = (\mathbf{H} \cdot \mathbf{C}^T) + \mathbf{\alpha} \odot \mathbf{z}$};
\node (linear) [above=of elementwise_mult, draw, fill=green!20, minimum height=0.9cm] {Linear Layer};
\node (Tabular head) [above=of input, yshift=7.8cm, draw, fill=red!20, minimum width=5.5cm,] {Tabular head};

\begin{scope}[on background layer]
  \fill[gray!10, rounded corners, draw=black, very thick] ($(hidden_update.north west) + (-0.75, 3.0)$) rectangle ($(linear.south east) + (3.9, -5.2)$);
\end{scope}
\draw [decorate, decoration={brace, amplitude=10pt, raise=5pt}, thick] 
  ($(hidden_update.north east) + (3.9, 3.0)$) -- 
  ($(linear.south east) + (3.9, -5.2)$) 
  node[midway, xshift=0.8cm, yshift=-0.2cm, rotate=90, draw=none, minimum width=0cm, minimum height=0cm] {$N \times$ Mamba Block};

\draw[arrow] (input) --  (embedding);
\draw[arrow] (encoding) -- ++(-2cm,0);
\draw[arrow] (embedding) -- (conv);
\draw[arrow] (conv) -- (x_part);
\draw[arrow, bend right] (embedding) to (z_part);
\draw[arrow] (x_part) -- (hidden_update);
\draw[arrow] (hidden_update) --  (elementwise_mult);
\draw[arrow] (elementwise_mult) -- (linear);
\draw[arrow] (D-matt) -- (elementwise_mult.east);
\draw[arrow] (hidden_update.west) -- (ssm.south);
\begin{scope}[on background layer]
  \draw[arrow, dashed] (z_part) -- (linear.east);
\end{scope}

\draw[arrow, bend right] (A-matt) to (hidden_update);
\draw[arrow] (x_part) to (A-matt);
\draw[arrow] (linear) to (Tabular head);

\end{tikzpicture}

\caption{The forward pass of a single sequence in the model. After embedding the inputs, the embeddings are passed to several Mamba blocks. The tabular head is a single task specific output layer. Before being passed to the Linear Layer, the contextualized embeddings are pooled via average pooling. For bidirectional processing a second block with a flipped sequence is used and the learnable matrices are not shared between the directions.}
\label{fig:forward_pass}
\end{figure}
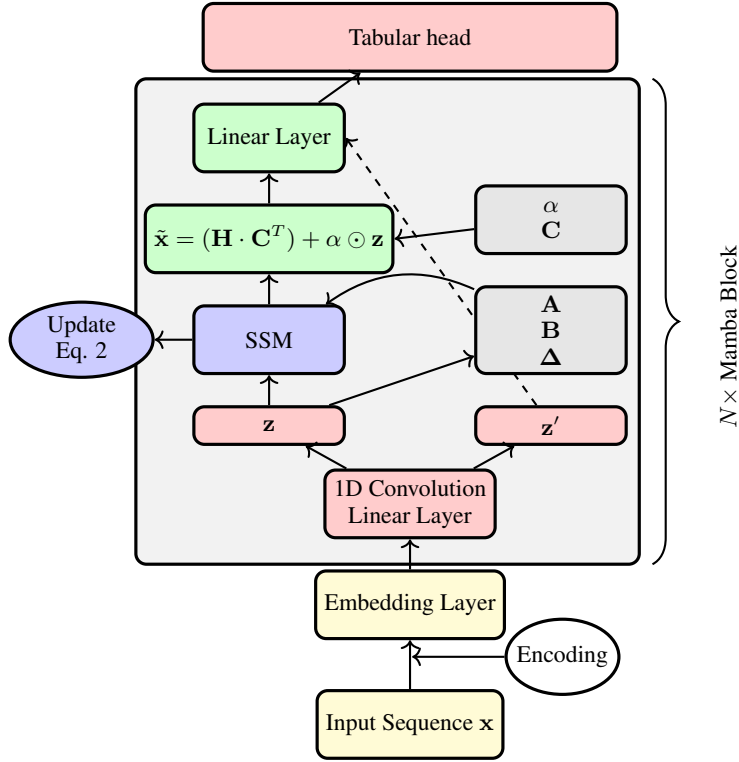

The model is trained end-to-end by minimizing the task-specific loss, e.g., mean squared error for regression or categorical cross entropy for classification tasks. An overview of a forward pass in the model is given in Figure \ref{fig:mamba_arch}.

\paragraph{MambAttention}
To accurately compare the introduced architecture with Transformer-based architectures, we also evaluate it against a model architecture, subsequently referred to as MambAttention, which is a simple combination of the proposed model and the FT-Transformer.
This architecture loosely follows the modeling approach from \cite{lieber2024jamba}. The input sequence is embedded as described previously, and the matrices $\mathbf{A}, \mathbf{B}, \Delta, \bar{\mathbf{z}}$ are constructed identically. However, after the first Mamba layer, a multi-head attention layer follows, similar to \cite{gorishniy2021revisiting}. The Mamba output representation, $\tilde{\mathbf{x}}_{\text{final}} \in \mathbb{R}^{N \times J \times d}$, is passed to the attention blocks in the same way as any embedded sequence representation.
Following \cite{huang2020tabtransformer}, no positional embeddings are used, and, in contrast to \citet{gorishniy2021revisiting}, no \textsc{[cls]} token is introduced to maintain consistent dimensions across all model blocks. Subsequently, Mamba and attention blocks are used in an alternating fashion, with both the first and last block always being a Mamba block. The final model representation is obtained similarly to the Mambular model, by pooling over the output representation sequence.

\section{Experiments}

\begin{figure*}
    \centering
    \includegraphics[width=\textwidth]{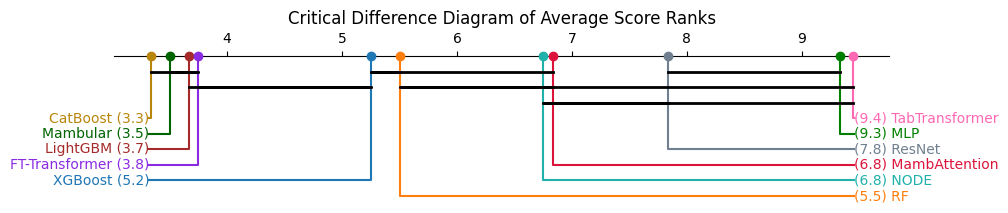}
    \caption{Critical difference diagram for all models on the benchmark datasets reported in Table \ref{tab:all_results}. The average ranks across tasks are shown in brackets next to each model. Horizontal lines indicate groups of models with no statistically significant differences in performance. \textit{Mambular} achieves the second-best average rank, with no significant differences at the 5\% level to the best performing model, \textit{CatBoost}. Notably, the top four models do not exhibit statistically significant differences from one another. The critical differences are computed using the Conover-Friedman test \citep{pereira2015overview}, as both average ranks and performance metrics across all datasets are available.}
    \label{fig:critical_diff_all}
\end{figure*}

\begin{figure*}[t]
    \centering
    \includegraphics[width=\textwidth]{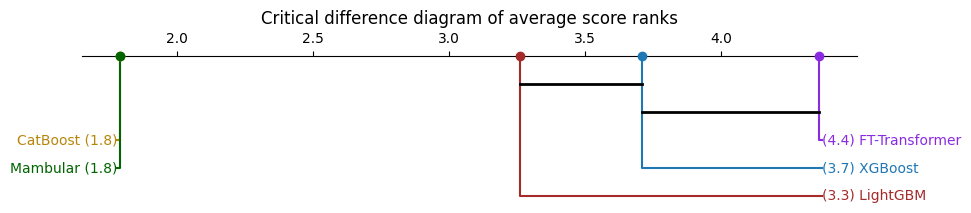}
    \caption{Critical difference diagram for best performing models on additional results reported in Table \ref{tab:add_results}. The average ranks across tasks are shown in brackets next to each model. Horizontal lines indicate groups of models with no statistically significant differences in performance. \textit{Mambular}  and \textit{CatBoost} achieve the best average (identical) rank and significantly outperform the other models.  The critical differences are computed using the Conover-Friedman test \citep{pereira2015overview}, as both average ranks and performance metrics across all datasets are available.}
    \label{fig:crit_difference_regression}
\end{figure*}

The presented autoregressive architectures are evaluated against a range of top-performing models \citep{mcelfresh2024neural} across multiple datasets (Supplementary Table~\ref{tab:datasets}). These models include FT-Transformer \citep{gorishniy2021revisiting}, TabTransformer \citep{huang2020tabtransformer}, XGBoost \citep{grinsztajn2022tree, mcelfresh2024neural}, LightGBM \citep{ke2017lightgbm}, CatBoost \citep{prokhorenkova2018catboost}, NODE \citep{popov2019neural}, a Random Forest, a baseline Multi-Layer Perceptron, and a ResNet. TabPFN \citep{hollmann2022tabpfn} is excluded due to its unsuitability for larger datasets. 

A 5-fold cross-validation is conducted for all datasets, with average results and standard deviations reported. PLE encodings (Eq. \ref{eq:ple}) with a maximum number of bins equal to the model dimension are used for all neural models (128 for most models, including MLP and ResNet). All categorical features are integer-encoded. For regression tasks, targets are normalized. Mean Squared Error (MSE) and Area Under the Curve (AUC) metrics are reported for regression and classification tasks respectively. TabTransformer, FT-Transformer, Mambular and MambAttention all employ identical architectures for embeddings and task-specific heads, which includes a single output layer without activation function or dropout. The \textsc{[cls]} token embedding is utilized for final prediction in the FT-Transformer as it has been shown to enhance performance \citep{thielmanninterpretable, gorishniy2021revisiting}.

All neural models share several parameters: a starting learning rate of 1e-04, weight decay of 1e-06, an early stopping patience of 15 epochs with respect to the validation loss, a maximum of 200 epochs for training, and learning rate decay with a factor of 0.1 with a patience of 10 epochs with respect to the validation loss. A universal batch size of 128 is used, and the best model with respect to the validation loss is returned for testing. TabTransformer, FT-Transformer, Mambular and MambAttention use the same embedding functions. For the benchmarks, a basic Mambular architecture is employed, using average pooling, no feature interaction layer, and no bi-directional processing. The columns/sequence are always sorted with numerical features first, followed by categorical features. Within these two groups, the features are sorted as they were originally provided in the dataset from the UCI Machine Learning Repository. A small kernel size of 4 in the convolutional layer is used based on the default Mamba architecture. The impact of variable positioning (with respect to the kernel size) on sequential processing is analyzed in section \ref{sec:seq_order}.
Details on the used datasets and preprocessing can be found in Appendix \ref{app:datasets}. Details on the model architectures and hyperparameters can be found in Appendix \ref{app:hparams}.

\begin{table*}[t]
\centering
\caption{Mean AUC and Mean MSE for various datasets and model configurations. We test different pooling methods, bi-directional processing and a learnable interaction layer. Significantly worse results compared to the default (average pooling, no interaction and no bi-directional processing) are marked red and bold at the 5\% significance level and underscored and red at the 10\% significance level. All results are achieved with 5-fold cross validation with identical seeds to the main results. \\}
\begin{tabular}{ccc|cc|cc}
\toprule
\textbf{Pooling} &  \textbf{bi-directional} & \textbf{Interaction} & \textbf{BA} $\uparrow$ & \textbf{AD} $\uparrow$ & \textbf{AB} $\downarrow$ & \textbf{CA} $\downarrow$\\
\midrule
Last            & $\times$        & $\times$        & \textbf{\textcolor{red}{0.916}} \std{0.004} & 0.927 \std{0.002} & 0.449 \std{0.043} & \textcolor{red}{\underline{0.181} \std{0.012}} \\
Sum             &  $\times$       & $\times$        & 0.925 \std{0.005} & 0.928 \std{0.002} & 0.449 \std{0.048} & 0.171 \std{0.009} \\
Max             &  $\times$       & $\times$        & 0.928 \std{0.004} & 0.927 \std{0.002} & 0.455 \std{0.050} & 0.172 \std{0.008} \\
\textsc{[cls]}  & $\times$        & $\times$        & \textbf{\textcolor{red}{0.914}} \std{0.005} & 0.928 \std{0.002} & 0.478 \std{0.044} & \textbf{\textcolor{red}{0.194}} \std{0.018} \\
\midrule
Avg             & $\checkmark$  & $\times$        & 0.927 \std{0.004} & 0.928 \std{0.002} & 0.450 \std{0.045} & 0.170 \std{0.010} \\
Avg             & $\times$        & $\checkmark$  & 0.928 \std{0.004} & 0.928 \std{0.002} & 0.453 \std{0.046} & 0.170 \std{0.007} \\
\midrule
Avg  & $\times$           & $\times$ & 0.927 \std{0.006} & 0.928 \std{0.002} & 0.452 \std{0.043} & 0.167 \std{0.011} \\
\bottomrule
\end{tabular}

\label{tab:results}
\end{table*}

\paragraph{Results}

Figure \ref{fig:critical_diff_all} provides a comprehensive ranking of all evaluated methods and their performance over all tasks. The results align with existing literature, highlighting the strong performance of the FT-Transformer architecture \citep{gorishniy2021revisiting}, LightGBM, CatBoost and XGBoost \citep{mcelfresh2024neural}. CatBoost emerges as the overall best-performing model across all tasks. Among the evaluated models, Mambular stands out as the top-performing neural model on average across all datasets and significantly outperforms XGBoost. Surprisingly, MambAttention, despite combining the second- and fourth-best models, underperforms compared to both. A potential reason is the disruption of information flow, which architectures like Jamba \citep{lieber2024jamba} counteract by structuring attention layers exclusively after six Mamba blocks and integrating a mixture of experts. However, while this approach enhances expressivity in larger models, it can lead to pronounced overfitting on these comparably smaller tabular datasets.

Detailed results for all datasets and tasks can be found in Table \ref{tab:all_results} in Appendix \ref{app:results}, with additional results on further models provided in Appendix \ref{app:add_results}. 


To further validate the potential of autoregressive models as an alternative for tabular deep learning, we compare the top five models on the benchmark suite from \cite{fischer2023openmlctr}. The results, presented in Figure \ref{fig:crit_difference_regression} and in detail in Appendix \ref{app:add_results} in Table \ref{tab:add_results} , show that Mambular and CatBoost significantly outperform the other models on average. We use the single train-test split provided by \cite{fischer2023openmlctr} and exclude datasets already listed in Table \ref{tab:all_results} to avoid redundancy.

\paragraph{Distributional Regression}

To validate Mambular's suitability for tabular problems, we conducted a small task on distributional regression \citep{Kneib}. Mambular for Location Scale and Shape (MambularLSS) outperforms XGBoostLSS \citep{marz2019xgboostlss} in terms of Continuous Ranked Probability Score (CRPS) \citep{gneiting2007strictly} when minimizing the negative log-likelihood while maintaining a small MSE. A detailed analysis can be found in Appendix \ref{app:lss}.

\begin{table}[ht]
    \centering
    \caption{Results for distributional regression for a normal distribution for the Abalone and California Housing datasets. Significantly better models at the 5\% level are marked in \textcolor{mygreen}{green}. $p$-vales are 0.20 and 0.00002 respectively for Abalone and and CA housing for the CRPS metric. \\}
 
    \begin{tabular}{c|cccc}
    \toprule
     & \multicolumn{2}{c}{\textbf{AB}} & \multicolumn{2}{c}{\textbf{CA}} \\
    & \textbf{CRPS} $\downarrow$ & \textbf{MSE} $\downarrow$ & \textbf{CRPS} $\downarrow$ & \textbf{MSE} $\downarrow$ \\
    \midrule
        MambularLSS & \textbf{0.345} \std{0.016} & \textbf{0.458} & \textbf{\textcolor{mygreen}{0.201}} \std{0.004} & \textbf{0.181} \\
        XGBoostLSS  & 0.359 \std{0.016} & 0.479 & 0.227 \std{0.005} & 0.215 \\
        \bottomrule
    \end{tabular}
    
    \label{tab:lss_results}
\end{table}

\section{Ablation}\label{seq:ablation}
\label{sec:ablation}
\paragraph{Model Architecture} 
This section explores the impact of various elements of Mambular's architecture, including (i) different pooling techniques, (ii) interaction layers, and (iii) bidirectional processing (Table~\ref{tab:results}). Transformer networks for natural language processing often use \textsc{[cls]} token embeddings for pooling \citep{gorishniy2021revisiting}, a technique that has also proven beneficial in tabular problems \citep{thielmanninterpretable}. Therefore, this technique is evaluated here. For pooling techniques, we compared Sum-pooling, Max-pooling, Last token pooling -- where only the last token in the sequence is passed to the task-specific model head --, and \textsc{[cls]} pooling\footnote{Note that \textsc{[cls]} token is appended to the end of each sequence in this implementation.} against standard Average-pooling.  

Given the significance of feature interactions in tabular problems, we also assessed the effectiveness of a learnable interaction layer between the features. This layer learns an interaction matrix \( \mathbf{W} \in \mathbb{R}^{J\times J} \), such that \( \text{interactions} = \mathbf{zW} \), where \( \mathbf{z} \) is the input feature matrix, before being passed through the SSM. This evaluation was only implemented for the standard Average pooling technique.

\begin{table*}[t]
\centering
\caption{Analysis of results for CA Housing. Significantly worse results than the default ordering - numerical features: categorical features - and a kernel size of 4, are marked in red. Increasing the kernel size induces positional invariance for features within the sequence. \\}
 \resizebox{0.8\textwidth}{!}{%
\begin{tabular}{l|ccc}
\toprule
Model &  \textbf{Kernel=4} $\downarrow$ &\textbf{Kernel=J}& Ordering\\
\midrule
Num$\mid$Cat  & 0.167 \std{0.011} & - & {[}\textcolor{lavender}{\textbf{LO}}, \textcolor{lavender}{\textbf{LA}}, MA, TR, TB, Po, Hh, MI, OP{]}\\
Cat$\mid$Num  & 0.158 \std{0.007} & - & {[}OP, MI, Hh, Po, TB, TR, MA, \textcolor{lavender}{\textbf{LA}}, \textcolor{lavender}{\textbf{LO}}{]}\\
\hline
& 0.177 \std{0.007}                             &0.160 \std{0.007}& {[}\textcolor{lavender}{\textbf{LO}}, MA, \textcolor{lavender}{\textbf{LA}}, TR, TB, Po, Hh, MI, OP{]} \\
& 0.175 \std{0.008}                             &0.173 \std{0.009}& {[}\textcolor{lavender}{\textbf{LO}}, MA, TR, \textcolor{lavender}{\textbf{LA}}, TB, Po, Hh, MI, OP{]} \\
& \textbf{\textcolor{red}{0.194}} \std{0.010}   &0.169 \std{0.008}& {[}\textcolor{lavender}{\textbf{LO}}, MA, TR, TB, \textcolor{lavender}{\textbf{LA}}, Po, Hh, MI, OP{]} \\
& \textbf{\textcolor{red}{0.196}} \std{0.011}   &0.161 \std{0.012}& {[}\textcolor{lavender}{\textbf{LO}}, MA, TR, TB, Po, \textcolor{lavender}{\textbf{LA}}, Hh, MI, OP{]} \\
& \textbf{\textcolor{red}{0.194}} \std{0.011}   &0.173 \std{0.009}& {[}\textcolor{lavender}{\textbf{LO}}, MA, TR, TB, Po, Hh, \textcolor{lavender}{\textbf{LA}}, MI, OP{]} \\
& \textbf{\textcolor{red}{0.195}} \std{0.010}   &0.169 \std{0.009}& {[}\textcolor{lavender}{\textbf{LO}}, MA, TR, TB, Po, Hh, MI, \textcolor{lavender}{\textbf{LA}}, OP{]} \\
& \textbf{\textcolor{red}{0.194}} \std{0.012}   &0.172 \std{0.011}& {[}\textcolor{lavender}{\textbf{LO}}, MA, TR, TB, Po, Hh, MI, OP, \textcolor{lavender}{\textbf{LA}}{]} \\

\bottomrule
\end{tabular}
}
\label{tab:ca_mambular_large_kernel}
\end{table*}

Interestingly, none of the configurations outperformed the basic architecture of average pooling, no interaction, and one-directional processing. Among the pooling strategies, last token pooling and \textsc{[cls]} token pooling performed significantly worse on two out of the four tested datasets. For this ablation study, a 5-fold cross-validation was performed, with the same hyperparameters used across all models. In bi-directional processing, each direction has its own set of learnable parameters, meaning that bi-directional models have additional trainable parameters. All model configurations can be found in Appendix \ref{app:hparams}.

\paragraph{Sequence ordering} \label{sec:seq_order}

Unlike models that leverage attention layers, Mambular is a sequential model. However, tabular data is not inherently sequential -- i.e., the order of features in tabular datasets should not matter. Therefore, we examined the significance of variables' position within the sequence and how their order impacts model performance. 

\begin{table}[h]
\centering
\caption{Mean AUC and Mean MSE for different feature orderings in the sequence. Flipping. the sequence does not significantly affect the performance at the 5\% or 10\% significance level. Significantly different values at the 5\% level from the default configuration (Num$\mid$Cat) are in bold and marked \textcolor{red}{\textbf{red}}. \\}

\begin{tabular}{l|cc|cc}
\toprule
Model & \textbf{BA} $\uparrow$ & \textbf{AD} $\uparrow$ & \textbf{AB} $\downarrow$ & \textbf{CA} $\downarrow$\\
\midrule
Num$\mid$Cat & 0.927 \std{0.006} & 0.928 \std{0.002} & 0.452 \std{0.043} & 0.167 \std{0.011} \\
Cat$\mid$Num & 0.925 \std{0.004} & 0.927 \std{0.002} & 0.454 \std{0.044} & 0.158 \std{0.007}\\
\hline 
random shuffle &  0.923 \std{0.002} & 0.927 \std{0.002} & 0.457 \std{0.045} & 0.172 \std{0.070}\\
random shuffle &  0.921 \std{0.005} & 0.927 \std{0.002} & 0.459 \std{0.049} & 0.177 \std{0.010}\\
random shuffle &  0.924 \std{0.005} & 0.927 \std{0.002} & 0.453 \std{0.045} & \textcolor{red}{\textbf{0.190}} \std{0.010}\\
\bottomrule
\end{tabular}

\label{tab:results_ordering}
\end{table}

In textual data, shuffling the order of words/tokens significantly affects the outcome, and even swapping single words can lead to entirely different contextualized embeddings. Since these contextualized representations are pooled and fed directly to Mambular's task-specific head, this could also impact performance.

Evaluation experiments were conducted on four real-world datasets and simulated data (see Appendix \ref{app:sequence_order}). As illustrated in Table~\ref{tab:results_ordering}, we initially confirmed the impact of the kernel size on tabular problems using Mamba's default kernel size of 4. The findings indicate that the order of sequences does not significantly influence model performance at the 5\% level for the selected datasets, even with a relatively small kernel size. However, this is contingent on the data. Strong interaction effects between features that are positioned further apart than the kernel size in the pseudo-sequence can negatively impact model performance, as demonstrated by the results on the California housing dataset.

The positions of the variables \textit{Longitude} and \textit{Latitude} appear to directly affect model performance (Table~\ref{tab:ca_mambular_large_kernel}). Performance begins to decline significantly when \textit{Longitude} and \textit{Latitude} are outside the kernel window. This issue can be entirely resolved by increasing the kernel size to match the sequence length $J$. For a comprehensive analysis, refer to Appendix \ref{app:sequence_order}.

\section{Limitations}

The model we have presented has been tested across various datasets and compared against a range of models. However, we have not conducted hyperparameter tuning, as findings from \citet{grinsztajn2022tree} and \citet{gorishniy2021revisiting} suggest that most models perform adequately without tuning. These studies indicate that while hyperparameter tuning can enhance performance across all models simultaneously, it does not significantly alter the relative ranking of the models. This suggests that a model that performs best or worst with default configurations will likely retain its ranking even after extensive tuning. Furthermore, \citet{mcelfresh2024neural} reported similar findings, strengthening the notion that hyperparameter tuning benefits most models equally without changing their comparative performance.

The absence of tuning does leave potential for enhancement across all models. However, the default configurations for the comparison models have been extensively tested in numerous studies. It is anticipated that if any model could gain more from hyperparameter tuning, it would be Mambular, due to the lack of extensive literature guiding its default settings. For the comparison models, we made our selections based on literature to ensure default parameters that are meaningful and high-performing. We managed to replicate average results from studies such as \citet{gorishniy2021revisiting} and \citet{grinsztajn2022tree}. Moreover, key hyperparameters like learning rate, patience, and number of epochs are shared among all models for a more uniform approach. All hyperparameter configurations can be found in Appendix \ref{app:hparams}.

\section{Conclusion}
We revisited autoregressive models for tabular deep learning and tested the proposed Mambular and MambAttention architectures against common benchmark models. Our work demonstrates the applicability of autoregressive architectures to tabular problems, offering a unique perspective on the interpretation and management of tabular data by treating it as a sequence. Interestingly, our findings suggest that the combination of autoregressive models and attention blocks significantly underperforms their parent models, which rely solely on autoregression or attention.
The performance of Mambular, along with its extension MambularLSS, highlights its broad applicability to a wide range of tabular tasks. Additionally, investigating the optimal feature ordering or integrating column-specific information through textual embeddings could further enhance performance.
Viewing tabular data as a sequence offers significant benefits for feature-incremental learning: new features can be directly appended to the sequence, eliminating the need to retrain the entire model. While Mamba is still relatively new compared to architectures like the Transformer, its rapid adoption suggests substantial potential for further advancement.

\clearpage
\bibliography{bib}
\bibliographystyle{apalike}

\newpage
\appendix

\section{Results} \label{app:results}
Detailed model performances are given below.

\begin{table}[ht]
\centering
\caption{Benchmarking results for regression and binary classification tasks.  Average mean squared error values and area under the curve values over 5 folds and the corresponding standard deviations are reported. Smaller values are better for the regression tasks and larger values are better for the classification tasks. The best performing model is marked in bold. An overview over all datasets for this benchmark is given in the appendix \ref{app:datasets}. The critical difference diagram is shown in Figure \ref{fig:critical_diff_all}. 
}
\resizebox{\textwidth}{!}{
\begin{tabular}{l|ccccccc|ccccc}
\toprule
  &  \multicolumn{7}{c}{Regression Tasks} & \multicolumn{5}{c}{Classification Tasks} \\
Models  & DI $\downarrow$ & AB $\downarrow$ & CA $\downarrow$ & WI $\downarrow$ & PA $\downarrow$ & HS $\downarrow$ & CP $\downarrow$ & BA $\uparrow$ & AD $\uparrow$ & CH $\uparrow$ & FI $\uparrow$ & MA $\uparrow$ \\
\midrule
XGBoost         & 0.019 & 0.506 & 0.171  & 0.528  & 0.036  & 0.119  & 0.024 & 0.928  & \textbf{0.929} & 0.845  & 0.774  & 0.922  \\
                & \std{0.000} & \std{0.044}  & \std{0.007} & \std{0.008} & \std{0.004} & \std{0.024} & \std{0.004} & \std{0.004} & \std{0.002}  & \std{0.008} & \std{0.009} & \std{0.002} \\

RF              & 0.019  & 0.461  & 0.183  & \textbf{0.485}  & 0.028  & 0.121  & 0.025  & 0.923  & 0.896  & 0.851  & 0.789  & 0.917  \\
                & \std{0.001} & \std{0.052} & \std{0.008} & \std{0.007} & \std{0.006} & \std{0.018} & \std{0.002} & \std{0.006} & \std{0.002} & \std{0.008} & \std{0.012} & \std{0.004} \\

LightGBM        & 0.019  & 0.459  & 0.171  & 0.542  & 0.039  & 0.112 & 0.023  & \textbf{0.932}  & 0.929  & 0.861  & 0.788  & \textbf{0.927}  \\
                & \std{0.001} & \std{0.047} & \std{0.007} & \std{0.013} & \std{0.007} & \std{0.018}  & \std{0.003} & \std{0.004} & \std{0.001} & \std{0.008} & \std{0.010} & \std{0.001} \\

CatBoost        &  0.019   & 0.457  & 0.169  & 0.583 & 0.045 & \textbf{0.106}  & \textbf{0.022}  & 0.932  & 0.927   & \textbf{0.867}  &0.796  &   0.926  \\
                & \std{0.000} & \std{0.007} & \std{0.006} & \std{0.006} & \std{0.006} & \std{0.015} & \std{0.001} & \std{0.008} & \std{0.002} & \std{0.006} & \std{0.010} & \std{0.005} \\
\hline
FT-Transformer  & 0.018 & 0.458 & 0.169 & 0.615 & \textbf{0.024} & 0.111 & 0.024 & 0.926 & 0.926 & 0.863 & 0.792 & 0.916 \\
                 & \std{0.001} & \std{0.055} & \std{0.006} & \std{0.012} & \std{0.005} & \std{0.014} & \std{0.001} & \std{0.003} & \std{0.002} & \std{0.007} & \std{0.011} & \std{0.003} \\
                 
MLP             & 0.066 & 0.462 & 0.198 & 0.654 & 0.764 & 0.147 & 0.031 & 0.895 & 0.914 & 0.840 & 0.793 & 0.886\\
                 & \std{0.003} & \std{0.051} & \std{0.011} & \std{0.013} & \std{0.023} & \std{0.017} & \std{0.001} & \std{0.004} & \std{0.002} & \std{0.005} & \std{0.011} & \std{0.003} \\
                 

ResNet          & 0.039 & 0.455 & 0.178 & 0.639 & 0.606 & 0.141 & 0.030 & 0.896 & 0.917 & 0.841 & 0.793 & 0.889\\
                & \std{0.018} & \std{0.045} & \std{0.006} & \std{0.013} & \std{0.031} & \std{0.017} & \std{0.002} & \std{0.006} & \std{0.002} & \std{0.006} & \std{0.013} & \std{0.003} \\
                 
NODE            & 0.019  &  \textbf{0.431}  & 0.207 & 0.613   & 0.045   & 0.124   & 0.026  &  0.914  & 0.904  & 0.851  & 0.790  & 0.904  \\
                & \std{0.000} & \std{0.052} & \std{0.001}  & \std{0.006} & \std{0.007}  & \std{0.015} & \std{0.001} & \std{0.008} & \std{0.002} & \std{0.006} & \std{0.010}  &  \std{0.005} \\
\hline

Mambular        & \textbf{0.018}  & 0.452  & \textbf{0.167}  & 0.628  & 0.035  & 0.132  & 0.025 &   0.927  & 0.928  & 0.861  & \textbf{0.796}  & 0.917  \\
                & \std{0.000} & \std{0.043} & \std{0.011} & \std{0.010} & \std{0.005} & \std{0.020} & \std{0.002} & \std{0.006} & \std{0.002} & \std{0.008} & \std{0.013} &\std{0.003} \\
MambAttention   & 0.018 & 0.484 & 0.189 & 0.638 & 0.030 & 0.142 & 0.026 & 0.919 & 0.921 & 0.857 & 0.781 & 0.911 \\
                 & \std{0.000} & \std{0.052} & \std{0.006} & \std{0.003} & \std{0.006} & \std{0.024} & \std{0.002} & \std{0.004} & \std{0.002} & \std{0.004} & \std{0.009} & \std{0.001} \\
\bottomrule
\end{tabular}}
\label{tab:all_results}
\end{table}

Further results on a regression benchmark with a single train-test-validation split are reported below. The datasets are taken from \cite{fischer2023openmlctr} with datasets already present in the main results excluded.

\begin{table*}[ht]
    \centering
    \caption{Benchmarking results for the top-5 performing models for regression tasks from \citep{fischer2023openmlctr} using a single provided train-test split. The reported values correspond to mean squared error (MSE), where lower values indicate better performance. \textit{Mambular} and \textit{CatBoost} achieve the best results, with identical average ranks. The corresponding critical difference diagram is presented in Figure \ref{fig:crit_difference_regression}. \\}
    \resizebox{\textwidth}{!}{
    \begin{tabular}{l|cccccccccccccc|c}
        \toprule
        \textbf{Model} & \textbf{BH ↓} & \textbf{CW ↓} & \textbf{FF ↓} & \textbf{GS ↓} & \textbf{HI ↓} & \textbf{K8 ↓} & \textbf{AV ↓} & \textbf{KC ↓} & \textbf{MH ↓} & \textbf{NP ↓} & \textbf{PP ↓} & \textbf{SA ↓} & \textbf{SG ↓} & \textbf{VT ↓} & \textbf{Rank ↓} \\
        \midrule
        Mambular       & \textbf{0.021} & \textbf{0.701} & 0.272 & 0.057 & \textbf{0.595} & 0.168 & 0.018 & 0.137 & 0.085 & 0.003 & 0.402 & \textbf{0.015} & 0.318 & \textbf{0.003} & \textbf{1.79} \\
        FT-Transformer & 0.028 & 0.701 & 0.301 & 0.205 & 0.609 & 0.451 & 0.089 & 0.149 & 0.101 & 0.009 & 0.542 & 0.033 & 0.360 & 0.045 & 4.36 \\
        CatBoost       & 0.032 & 0.702 & \textbf{0.245} & \textbf{0.041} & 0.597 & \textbf{0.150} & 0.004 & \textbf{0.110} & \textbf{0.078} & 0.005 & \textbf{0.390} & 0.018 & \textbf{0.297} & 0.013 & \textbf{1.79} \\
        LightGBM       & 0.048 & 0.707 & 0.263 & 0.059 & 0.599 & 0.239 & 0.024 & 0.140 & 0.091 & 0.009 & 0.452 & 0.031 & 0.302 & 0.013 & 3.26 \\
        XGBoost        & 0.039 & 0.752 & 0.281 & 0.078 & 0.635 & 0.259 & \textbf{0.004} & 0.161 & 0.098 & 0.006 & 0.403 & 0.024 & 0.329 & 0.013 & 3.71 \\
        \bottomrule
    \end{tabular}
    }
    \label{tab:add_results}
\end{table*}

\section{Datasets}\label{app:datasets}
All used datasets are taken from the UCI Machine Learning repository and publicly available. We drop out all missing values. For the regression tasks we standard normalize the targets.
Otherwise, preprocessing is performed as described above. Note, that before PLE encoding we scale the numerical features to be within (-1, +1).

\begin{table}[ht]
    \centering
    \caption{The used datasets for benchmarking. All datasets are taken from the UCI Machine Learning repository. \#num and \#cat represent the number of numerical and categorical features respectively. The number of features thus determines for Mambular the "sequence length". The train, test and validation numbers represent the average number of samples in the respective split for the 5-fold cross validation. Ratio marks the percentage of the dominant class for the binary classification tasks.}
    \begin{tabular}{l|ccccccc}
    \toprule
        Name & Abbr. & \#cat & \#num & train & test & val & ratio\\
        \hline
        & \multicolumn{7}{c}{Regression Datasets}\\
        \hline
         Diamonds           & DI  & 4	&7	&34522	&10788	&8630 &-\\
         Abalone            & AB  & 1	&8	&2673	&835	&668&-\\
         California Housing & CA  & 1	&9	&13210	&4128	&3302&-\\
         Wine Quality       & WI  & 0	&12	&4158	&1299	&1039&-\\
         Parkinsons         & PA  & 2	&20	&3760	&1175	&940&-\\
         House Sales        & HS  & 8	&19	&13832	&4322	&3458&-\\
         CPU small          & CPU & 0	&13	&5243	&1638	&1310 &-\\
        \midrule
        & \multicolumn{7}{c}{Classification Datasets}\\
        \hline
        Bank        & BA & 13	&8	&28935	&9042	&7233 & 88.3\%\\
        Adult       & AD & 9	&6	&31259	&9768	&7814 & 76.1\%\\
        Churn       & CH & 3	&9	&6400	&2000	&1600 & 79.6\%\\
        FICO        & FI & 0	&32	&6694	&2091	&1673 &53.3\%\\
        Marketing   & MA & 15	&8	&27644	&8638	&6910 &88.4\%\\	
        \bottomrule
    \end{tabular}
    
    \label{tab:datasets}
\end{table}

\section{Sequence ordering} \label{app:sequence_order}
We test two different shuffling settings: \textbf{i)} shuffling the embeddings after they have passed through the embedding layer, \textbf{ii)} shuffling the sequence of variables before being passed through the embedding layers. 

All sequences are ordered by default with numerical features first, followed by categorical features, as arranged in the datasets from the UCI Machine Learning Repository. For the ablation study, a dataset with 5,000 samples and 10 features—five numerical and five categorical—was simulated. The numerical features were generated with large correlations, including two pairs with correlations of 0.8 and 0.6, respectively. The categorical features were created with four distinct categories.
Interaction terms were included as follows: An interaction between two numerical features, an interaction between a categorical and a numerical feature, and an interaction between two categorical features. The numerical features were scaled using standard normalization before generating the target variable. The target variable was constructed to include linear effects from each feature and the specified interaction terms, with added Gaussian noise for variability.  We first fit a XGBoost model  for a sanity check. Subsequently, we fit Mambular with default ordering (numerical before categorical features), flipped ordering and switched categorical and numerical ordering. Subsequently, we randomly shuffled the order and fit 10 models. We find that ordering does not have an effect on this simulated data, even with these large interaction and correlation effects\footnote{See the appendix for the chosen model parameters. Since the dataset is comparably smaller, we used a smaller Mambular model. Hyperparameters such as the learning rate, batch size etc. are kept identical to the default Mambular model.}.

\begin{table}[ht]
\centering
\caption{Performance for different orderings of features. Numerical features are given as integer numbers, categorical features as capital letters. Feature interaction between numerical features is given in \textcolor{blue}{blue}. Feature interaction between categorical features is denoted in \textcolor{mygreen}{green} and feature interaction between a numerical and a categorical feature is given in \textcolor{lavender}{lavender}. We find that reordering the features either before or after the embedding layers does not affect performance of the model. No ordering performs significantly better or worse than the default model, while all models perform significantly better than the XGBoost model.}
\begin{tabular}{ccc}
\toprule
 \textbf{Before Embedding Layer} & \textbf{After Embedding Layer} & \textbf{Ordering} \\ 
\midrule
 Default & 0.918 \std{0.045}  &{[}\textcolor{blue}{1 2} \textcolor{lavender}{3} 4 5 \textcolor{lavender}{A} \textcolor{mygreen}{B C}  D E{]} \\ 
\hline
 0.916 \std{0.043} & 0.913 \std{0.043} &{[}E D \textcolor{mygreen}{C B} \textcolor{lavender}{A} 5 4 \textcolor{lavender}{3} \textcolor{blue}{2 1}{]} \\ 
 0.919 \std{0.044} & 0.914 \std{0.042} &{[}\textcolor{lavender}{A} \textcolor{mygreen}{B C}  D E \textcolor{blue}{1 2} \textcolor{lavender}{3} 4 5{]} \\
\hline
0.917 \std{0.043} & 0.915 \std{0.045} &{[}\textcolor{lavender}{A} \textcolor{mygreen}{B} \textcolor{blue}{2} \textcolor{lavender}{3} \textcolor{blue}{1} D E 4 \textcolor{mygreen}{C} 5{]} \\ 
0.920\std{0.046} & 0.917 \std{0.045} &{[}D \textcolor{mygreen}{C} \textcolor{blue}{2} \textcolor{lavender}{A} \textcolor{mygreen}{B} E \textcolor{blue}{1} 5 \textcolor{lavender}{3} 4{]} \\ 
0.914 \std{0.043} & 0.914 \std{0.044} &{[}\textcolor{mygreen}{B} \textcolor{blue}{1} 4 \textcolor{mygreen}{C} D \textcolor{lavender}{A} \textcolor{blue}{2} E \textcolor{lavender}{3} 5{]} \\ 
0.916 \std{0.045} & 0.914 \std{0.041} &{[}\textcolor{blue}{1} 5 E \textcolor{mygreen}{B} \textcolor{mygreen}{C} 4 \textcolor{lavender}{3} D \textcolor{blue}{2} \textcolor{lavender}{A}{]} \\ 
0.918 \std{0.046} & 0.914 \std{0.045} &{[}\textcolor{blue}{2} 5 E \textcolor{mygreen}{B} 4 \textcolor{lavender}{A} \textcolor{blue}{1} \textcolor{lavender}{3} D \textcolor{mygreen}{C}{]} \\ 
0.916 \std{0.044} & 0.915 \std{0.043} &{[}\textcolor{blue}{1} \textcolor{mygreen}{C} \textcolor{lavender}{A} \textcolor{blue}{2} D 4 E \textcolor{lavender}{3} 5 \textcolor{mygreen}{B}{]} \\ 
0.917 \std{0.040} & 0.914 \std{0.043} &{[}\textcolor{lavender}{A} \textcolor{blue}{1} 4 5 \textcolor{blue}{2} \textcolor{mygreen}{C} E \textcolor{mygreen}{B} D \textcolor{lavender}{3}{]} \\ 
0.917 \std{0.044} & 0.922 \std{0.040} &{[}4 \textcolor{lavender}{A} \textcolor{blue}{1} \textcolor{blue}{2} \textcolor{lavender}{3} \textcolor{mygreen}{B} 5 \textcolor{mygreen}{C} D E{]} \\ 
0.920 \std{0.040} & 0.913 \std{0.040} &{[}\textcolor{blue}{1} \textcolor{lavender}{A} D \textcolor{mygreen}{C} \textcolor{mygreen}{B} \textcolor{lavender}{3} E \textcolor{blue}{2} 5 4{]} \\ 
0.920 \std{0.041} & 0.916 \std{0.044} &{[}\textcolor{mygreen}{C} 5 \textcolor{mygreen}{B} \textcolor{blue}{2} 4 \textcolor{lavender}{A} E D \textcolor{lavender}{3} \textcolor{blue}{1}{]} \\ 
\midrule
XGBoost & 1.096 \std{0.038}  & \\
\bottomrule
\end{tabular}

\label{table:performance_metrics}
\end{table}

\newpage
\subsection{California Housing}\label{app:housing}

The p-values for the sequence ordering and positioning of Latitude and Longitude is given below.

\begin{table}[ht]
\centering
\caption{Detailed analysis of results for CA Housing, including p-statistics.}
 
\begin{tabular}{l|ccc}
\toprule
Model &  \textbf{CA} $\downarrow$ &$p$-value& Ordering\\
\midrule
Num$\mid$Cat  & 0.167 \std{0.011} & - & {[}\textcolor{lavender}{\textbf{LO}}, \textcolor{lavender}{\textbf{LA}}, MA, TR, TB, Po, Hh, MI, OP{]}\\
Cat$\mid$Num  & 0.158 \std{0.007} &0.168& {[}OP, MI, Hh, Po, TB, TR, MA, \textcolor{lavender}{\textbf{LA}}, \textcolor{lavender}{\textbf{LO}}{]}\\
\hline
& 0.177 \std{0.007}                             &0.136& {[}\textcolor{lavender}{\textbf{LO}}, MA, \textcolor{lavender}{\textbf{LA}}, TR, TB, Po, Hh, MI, OP{]} \\
& 0.175 \std{0.008}                             &0.240& {[}\textcolor{lavender}{\textbf{LO}}, MA, TR, \textcolor{lavender}{\textbf{LA}}, TB, Po, Hh, MI, OP{]} \\
& \textbf{\textcolor{red}{0.194}} \std{0.010}   &0.003& {[}\textcolor{lavender}{\textbf{LO}}, MA, TR, TB, \textcolor{lavender}{\textbf{LA}}, Po, Hh, MI, OP{]} \\
& \textbf{\textcolor{red}{0.196}} \std{0.011}   &0.003& {[}\textcolor{lavender}{\textbf{LO}}, MA, TR, TB, Po, \textcolor{lavender}{\textbf{LA}}, Hh, MI, OP{]} \\
& \textbf{\textcolor{red}{0.194}} \std{0.011}   &0.004& {[}\textcolor{lavender}{\textbf{LO}}, MA, TR, TB, Po, Hh, \textcolor{lavender}{\textbf{LA}}, MI, OP{]} \\
& \textbf{\textcolor{red}{0.195}} \std{0.010}   &0.004& {[}\textcolor{lavender}{\textbf{LO}}, MA, TR, TB, Po, Hh, MI, \textcolor{lavender}{\textbf{LA}}, OP{]} \\
& \textbf{\textcolor{red}{0.194}} \std{0.012}   &0.005& {[}\textcolor{lavender}{\textbf{LO}}, MA, TR, TB, Po, Hh, MI, OP, \textcolor{lavender}{\textbf{LA}}{]} \\

\bottomrule
\end{tabular}

\label{tab:ca_mambular}
\end{table}

Given these results, and to verify, that the kernel size of 4 is the cause of this effect, we further analyzed the dataset.
Below are more results for Mambular with random shuffling. Again we can see the the position of Latitude and Longitude significantly impact model performance, whenever these two variables are further apart than the fixed kernel size of 4. 
\begin{table}[ht]
\centering
\caption{Analysis of results for CA Housing}
 
\begin{tabular}{l|ccc}
\toprule
Model &  \textbf{CA} $\downarrow$  & $p$-value & Ordering\\
\midrule
Num$\mid$Cat  & 0.167 \std{0.011} & -& {[}\textcolor{lavender}{\textbf{LO}}, \textcolor{lavender}{\textbf{LA}}, MA, TR, TB, Po, Hh, MI, OP{]} \\
Cat$\mid$Num  & 0.158 \std{0.007} & 0.168& {[}OP, MI, Hh, Po, TB, TR, MA, \textcolor{lavender}{\textbf{LA}}, \textcolor{lavender}{\textbf{LO}}{]} \\
\hline
& 0.174  \std{0.009} & 0.304 & {[}Po, Hh, MI, OP, \textcolor{lavender}{\textbf{LO}}, \textcolor{lavender}{\textbf{LA}}, MA, TR, TB{]} \\
& \textcolor{red}{\textbf{0.195}}  \std{0.012} &0.005 & {[}\textcolor{lavender}{\textbf{LO}}, MA, TR, TB, Po, Hh, MI, OP, \textcolor{lavender}{\textbf{LA}}{]} \\
& \textcolor{red}{\textbf{0.197}}  \std{0.010} &0.002 & {[}MA, \textcolor{lavender}{\textbf{LO}}, TR, TB, Po, Hh, MI, \textcolor{lavender}{\textbf{LA}}, OP{]} \\
& \textcolor{red}{\textbf{0.188}}  \std{0.010} &0.014 & {[}MA, TR, \textcolor{lavender}{\textbf{LO}}, TB, Po, Hh, \textcolor{lavender}{\textbf{LA}}, MI, OP{]} \\
& 0.178  \std{0.010} &0.137 & {[}MA, \textcolor{lavender}{\textbf{LO}}, \textcolor{lavender}{\textbf{LA}}, TR, TB, Po, Hh, MI, OP{]} \\
& 0.177  \std{0.008} &0.142 & {[}MA, TR, TB, Po, \textcolor{lavender}{\textbf{LA}}, \textcolor{lavender}{\textbf{LO}}, Hh, MI, OP{]} \\
& 0.178  \std{0.009} &0.123 & {[}\textcolor{lavender}{\textbf{LA}}, \textcolor{lavender}{\textbf{LO}}, MA, TR, TB, Po, Hh, MI, OP{]} \\
\hline

& 0.172 \std{0.070} &0.420 & {[}Hh, TB, Po, MI, MA, OP, \textcolor{lavender}{\textbf{LA}}, \textcolor{lavender}{\textbf{LO}}, TR{]}\\
& 0.177 \std{0.010} &0.171 & {[}\textcolor{lavender}{\textbf{LO}}, Po, OP, \textcolor{lavender}{\textbf{LA}}, MI, MA, TR, Hh, TB{]}\\
& \textcolor{red}{\textbf{0.190}} \std{0.010} &0.009 & {[}Hh, TB, \textcolor{lavender}{\textbf{LO}}, MI, Po, OP, TR, MA, \textcolor{lavender}{\textbf{LA}}{]}\\
\bottomrule
\end{tabular}

\label{tab:results_order}
\end{table}

To analyze the feature interaction effect between these two variables, we conducted a simple regression with pairwise feature interactions and analyzed the effect strengths. Interestingley, we find that the interaction between Longitude and Latitude is not as prominent as that between other variables.
\begin{figure}[ht]
    \centering
    \includegraphics[width=0.9\linewidth]{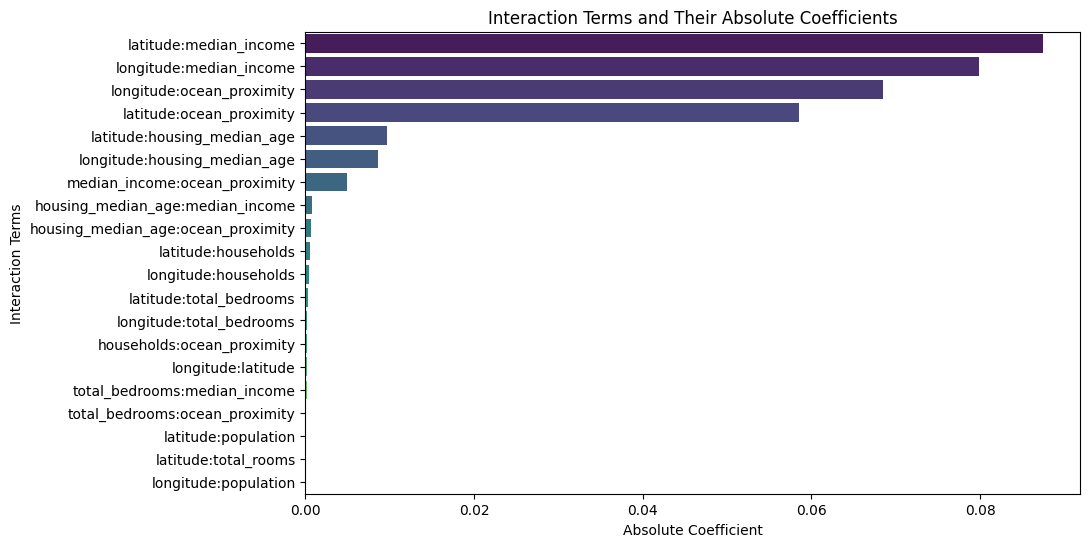}
    \caption{Linear Regression with pairwise interaction effects on the california housing dataset.}
    \label{fig:enter-label}
\end{figure}

Additionally, we have fit a XGboost model and analyzed the pairwise feature importance metrics and generally find the same results as for the linear regression.

\begin{figure}[ht]
    \centering
    \includegraphics[width=0.9\linewidth]{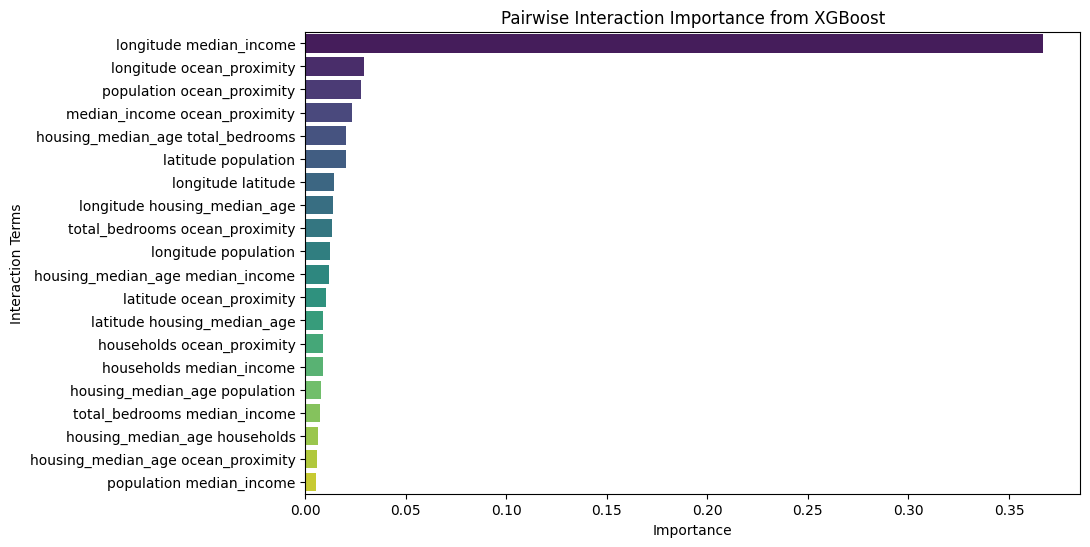}
    \caption{Pairwise feature importance statistics from a XGBoost model on the california housing dataset.}
    \label{fig:enter-label}
\end{figure}

\clearpage

\section{Distributional Regression} \label{app:lss}

Distributional regression describes regression beyond the mean, i.e., the modeling of all distributional parameters. Thus, Location Scale and Shape (LSS) models can quantify the effects of covariates on not just the mean but also on any parameter of a potentially complex distribution assumed for the responses. A major advantage of these models is their ability to identify changes in all aspects of the response distribution, such as variance, skewness, and tail probabilities, enabling the model to properly disentangling aleatoric uncertainty from epistemic uncertainty. 

This is achieved by minimizing the negative log-likelihood via optimizing the parameters $\theta$

\[ \mathcal{L}(\theta) = - \sum_{i=1}^n \log f(y_i \mid \mathbf{x}_i, \theta) \]

For the two examples in the main part, a normal distribution is modelled and hence, the models minimize:

\begin{equation*}
    log \left(\mathcal{L}(\mu, \sigma^2 | y)\right) = -\frac{n}{2}\log(2\pi\sigma^2) - \frac{1}{2\sigma^2}\sum_{i=1}^n(y_i - \mu)^2,
\end{equation*}\label{app:normal}
where $n$ is the underlying number of observations and parameters $y \in \mathbb{R}$, location $\mu \in \mathbb{R}$ and scale $\sigma \in \mathbb{R}^{+}$.

While this has been a common standard in classical statistical approaches \citep{stasinopoulos2008generalized}, it has not yet been widely adopted by the ML community. Recent interpretable approaches \citep{Thielmann}, however, have demonstrated the applicability of distributional regression in tabular deep learning. Furthermore, approaches like XGBoostLSS \citep{marz2019xgboostlss, marz2022distributional} demonstrate that tree-based models are capable of effectively solving such tasks. Below, we show that Mambular for Location Scale and Shape (MambularLSS) outperforms XGBoostLSS in terms of Continuous Ranked Probability Score (CRPS) \citep{gneiting2007strictly} when minimizing the negative log-likelihood while maintaining a small MSE. 

\paragraph{CRPS}\label{app:CRPS}

Analyzing distributional regression models also requires careful consideration of the evaluation metrics. Traditionally, mean focused models are evaluated using mean-centric metrics, e.g. MSE, AUC or Accuracy. However, a model that takes all distributional parameters into account should be evaluated on the predictive performance for all of the distributional parameters.
Following \cite{gneiting2007strictly}, the evaluation metric should be proper, i.e. enforce the analyst to report their true beliefs in terms of a predictive distribution. In terms of classical mean-centric metrics, e.g. MSE is proper for the mean, however, not proper for evaluating the complete distributional prediction.
We therefore rely on the Continuous Ranked Probability Score \citep{gneiting2007strictly} for model evaluation, given by:
$$CRPS(F, x) = -\int_{-\infty}^{\infty} (F(y) - \mathbf{1}_{y \geq x})^2 \, dy.$$ 
See \cite{gneiting2007strictly} for more details.

\begin{table}[ht]
    \centering
    \caption{Results for distributional regression for a normal distribution for the Abalone and California Housing datasets. Significantly better models at the 5\% level are marked in \textcolor{mygreen}{green}. $p$-vales are 0.20 and 0.00002 respectively for Abalone and and CA housing for the CRPS metric.}
    \begin{tabular}{c|cccc}
    \toprule
     & \multicolumn{2}{c}{\textbf{AB}} & \multicolumn{2}{c}{\textbf{CA}} \\
    & \textbf{CRPS} $\downarrow$ & \textbf{MSE} $\downarrow$ & \textbf{CRPS} $\downarrow$ & \textbf{MSE} $\downarrow$ \\
    \midrule
        MambularLSS & \textbf{0.345} \std{0.016} & \textbf{0.458} & \textbf{\textcolor{mygreen}{0.201}} \std{0.004} & \textbf{0.181} \\
        XGBoostLSS  & 0.359 \std{0.016} & 0.479 & 0.227 \std{0.005} & 0.215 \\
        \bottomrule
    \end{tabular}
    \label{tab:my_label}
\end{table}

\section{MambaTab}\label{app:mambatab}

In addition to the popular tabular models described above, we tested the architecture proposed by \citet{ahamed2024mambatab}. MambaTab is the first architecture to leverage Mamba blocks for tabular problems. However, the authors propose using a combined linear layer to project all inputs into a single feature representation, transforming the features into a pseudo-sequence of fixed length 1. This approach simplifies the recursive update from Eq. \ref{eq:ssm} into a matrix multiplication and makes the model resemble a ResNet due to the residual connections in the final processing. 
Utilizing a sequential model with a sequence length of 1 does not fully exploit the strengths of sequential processing, as it reduces the model's capacity to capture dependencies across multiple features.

We tested the architecture proposed by \citet{ahamed2024mambatab} and could achieve similar results for shared datasets, but overall found MambaTab to perform similar to a ResNet, aligning with expectations (see Table \ref{tab:averages} and \ref{tab:reg_regression}). Additionally, we experimented with transposing the axes to create an input matrix of shape \textsc{(1) $\times$ (Batch Size) $\times$ (Embedding Dimension)}, as outlined in their implementation. While this approach draws on ideas from TabPFN \citep{hollmann2022tabpfn}, it did not lead to performance improvements in our experiments. When using PLE encodings and increasing the number of layers and dimensions compared to the default implementation from \citet{ahamed2024mambatab} we are able to increase performance. 

MambaTab \citep{ahamed2024mambatab} significantly differs from Mambular, since it is not a sequential model. To achieve the presented results from MambaTab, we have followed the provided implementation from the authors retrieved from \url{https://github.com/Atik-Ahamed/MambaTab}. It is worth noting, however, that MambaTab benchmarks the model on a lot of smaller datasets. 50\% of the benchmarked datasets have not more than 1000 observations. Additionally, the provided implementation suggests, that MambaTab does indeed not iterate over a pseudo sequence length of 1, but rather over the number of observations, similar to a TabPFN \citep{hollmann2022tabpfn}. We have also tested this version, denoted as MambaTab$^T$ but did not find that it performs better than the described version. On the Adult dataset, our achieved result of 0.901 AUC on average is very similar to the default results reported in \citet{ahamed2024mambatab} with 0.906. The difference could be firstly due to us performing 5-fold cross validation and secondly different seeds in model initialization. 

\begin{table}[ht]
\centering
\caption{Benchmarking results for the regression tasks for the original MambaTab implementation provided by \url{https://github.com/Atik-Ahamed/MambaTab}}
\resizebox{\textwidth}{!}{%
\begin{tabular}{l|ccccccc}
\toprule
Models  & DI $\downarrow$ & AB $\downarrow$ & CA $\downarrow$ & WI $\downarrow$ & PA $\downarrow$ & HS $\downarrow$ & CP $\downarrow$ \\
\midrule
MambaTab & 0.035 \std{0.006} & 0.456 \std{0.053} & 0.272 \std{0.016} & 0.685 \std{0.015} & 0.531 \std{0.032} & 0.163 \std{0.009} & 0.030 \std{0.002} \\
MambaTab$^{T}$ & 0.038 \std{0.002} & 0.468 \std{0.048} & 0.279 \std{0.010} & 0.694 \std{0.015} & 0.576 \std{0.022} & 0.179 \std{0.027} & 0.033 \std{0.002} \\
\bottomrule
\end{tabular}
}

\label{tab:reg_regression-mambatab}
\end{table}

\begin{table}[ht]
\centering
\caption{Benchmarking results for the classification tasks.  Average AUC values over 5 folds and the corresponding standard deviations are reported. Larger values are better.}
\resizebox{\textwidth}{!}{%
\begin{tabular}{l|ccccc}
\toprule
Models & BA $\uparrow$ & AD $\uparrow$ & CH $\uparrow$ & FI $\uparrow$ & MA $\uparrow$\\
\midrule
MambaTab        & 0.886 \std{0.006} & 0.901 \std{0.001} & 0.828 \std{0.005} & 0.785 \std{0.012} & 0.880 \std{0.003} \\
MambaTab$^{T}$  & 0.888 \std{0.005} & 0.899 \std{0.002} & 0.815 \std{0.009} & 0.783 \std{0.012} & 0.878 \std{0.005} \\
\bottomrule
\end{tabular}
}

\label{tab:classification-results-mambatab}
\end{table}

\newpage
\section{Default Model Hyperparameters}\label{app:hparams}

In the following, we describe the default model parameters used for all the neural models. We based our choices on the literature to ensure meaningful and high-performing parameters by default. Additionally, we were able to reproduce results (on average) from popular studies, such as \citet{gorishniy2021revisiting} and \citet{grinsztajn2022tree}. While most larger benchmark studies perform extensive hyperparameter tuning for each dataset, analyzing these results \citep{grinsztajn2022tree, gorishniy2021revisiting} shows that most models already perform well without tuning, as also found by \citet{mcelfresh2024neural}. Furthermore, the results suggest that performing hyperparameter tuning for all models does not change the ranking between the models, since most models benefit from tuning to a similar degree. Thus, we have collected informed hyperparameter defaults which we list in the following. The hyperparameters such as learning rate, patience and number of epochs are shared among all models for a more consistent approach.

\begin{table}[ht]
\centering
\caption{Shared hyperparameters among all models}
\begin{tabular}{ll}
\toprule
\textbf{Hyperparameter} & \textbf{Value} \\
\midrule
Learning rate & 1e-04 \\
Learning rate patience & 10 \\
Weight decay & 1e-06 \\
Learning rate factor & 0.1 \\
Max Epochs & 200 \\
\bottomrule
\end{tabular}
\label{tab:shared_hyperparameters}
\end{table}

\paragraph{MLP} As a simple baseline,  we fit a simple MLP without any special architecture. However, PLE encodings are used, as they have been shown to significantly improve performance.

\begin{table}[ht]
\centering
\caption{Default Hyperparameters for the MLP Model}
\begin{tabular}{ll}
\toprule
\textbf{Hyperparameter} & \textbf{Value} \\
\midrule
Layer sizes & (256, 128, 32) \\
Activation function & SELU \\
Dropout rate & 0.5 \\
PLE encoding dimension & 128 \\
\bottomrule
\end{tabular}
\label{tab:mlp_hyperparameters}
\end{table}

\paragraph{ResNet} A ResNet architecture for tabular data has been shown to be a sensible baseline \citep{gorishniy2021revisiting}. Furthermore, \citet{mcelfresh2024neural} has validated the strong performance of ResNets compared to e.g. TabNet \citep{arik2021tabnet} or NODE \citep{popov2019neural}.

\begin{table}[ht]
\centering
\caption{Default Hyperparameters for the ResNet Model}
\begin{tabular}{ll}
\toprule
\textbf{Hyperparameter} & \textbf{Value} \\
\midrule
Layer sizes & (256, 128, 32) \\
Activation function & SELU \\
Dropout rate & 0.5 \\
Skip connections & True \\
Batch normalization & True \\
Number of blocks & 3 \\
PLE encoding dimension & 128 \\
\bottomrule
\end{tabular}
\label{tab:resnet_hyperparameters}
\end{table}

\paragraph{FT-Transformer} For the FT-Transformer architecture we orientated on the default parameters conducted by \citet{gorishniy2021revisiting}. We only slightly adapted them from 3 layers and an embedding dimension of 192 to 4 layers and an embedding dimension of 128 to be more consistent with the other models. However, we tested out the exact same architecture from \citet{gorishniy2021revisiting} and did not find any differences in performance, even a minimal (non-significant) decrease. Additionally, we found that using ReGLU instead of ReLU activation function in the transformer blocks does improve performance consistently.

\begin{table}[ht]
\centering
\caption{Default Hyperparameters for the FT Transformer Model}
\begin{tabular}{ll}
\toprule
\textbf{Hyperparameter} & \textbf{Value} \\
\midrule
Model Dim & 128 \\
Number of layers & 4 \\
Number of attention heads & 8 \\
Attention dropout rate & 0.2 \\
Feed-forward dropout rate & 0.1 \\
Normalization method & LayerNorm \\
Embedding activation function & Identity \\
Pooling method & cls \\
Normalization first in transformer block & False \\
Use bias in linear layers & True \\
Transformer activation function & ReGLU \\
Layer normalization epsilon & 1e-05 \\
Feed-forward layer dimensionality & 256 \\
PLE encoding dimension & 128 \\
\bottomrule
\end{tabular}
\label{tab:ft_transformer_hyperparameters}
\end{table}

\paragraph{TabTransformer} We practically used the same hyperparameter for TabTransformer as we used for Ft-Transformer. For consistency we do not use a multi-layer MLP for where the contextualized embeddings are being passed to. While this deviates from the original architecture, leaving this out ensures a more consistent comparison to FT-Transformer and Mambular since both models use a single layer after pooling. However, we used a larger feed forward dimensionality in the transformer to counteract this. Overall, our results are in line with the literature and we can validate that TabTransformer outperforms a simple MLP on average. For datasets where no categorical features are available, the TabTransformer converges to a simple MLP. Thus we left these results blank in the benchmarks.

\begin{table}[ht]
\centering
\caption{Default Hyperparameters for the TabTransformer Model}
\begin{tabular}{ll}
\toprule
\textbf{Hyperparameter} & \textbf{Value} \\
\midrule
Model Dim & 128 \\
Number of layers & 4 \\
Number of attention heads & 8 \\
Attention dropout rate & 0.2 \\
Feed-forward dropout rate & 0.1 \\
Normalization method & LayerNorm \\
Embedding activation function & Identity \\
Pooling method & cls \\
Normalization first in transformer block & False \\
Use bias in linear layers & True \\
Transformer activation function & ReGLU \\
Layer normalization epsilon & 1e-05 \\
Feed-forward layer dimensionality & 512 \\
PLE encoding dimension & 128 \\

\bottomrule
\end{tabular}
\label{tab:tabtransformer_hyperparameters}
\end{table}

\paragraph{MambaTab} We test out three different MambaTab architectures. Firstly, we implement the same architecture as for Mambular but instead of an embedding layer for each feature and creating a sequence of length $J$ we feed all features jointly through a single embedding layer and create a sequence of length 1. The \textit{Axis} argument thus specifies over which axis the SSM model iterates. As described by \citet{ahamed2024mambatab} the model iterates over this pseudo-sequence length of 1.

\begin{table}[ht]
\centering
\caption{Default Hyperparameters for the MambaTab$^*$ Model}
\begin{tabular}{ll}
\toprule
\textbf{Hyperparameter} & \textbf{Value} \\
\midrule
Model Dim & 64 \\
Number of layers & 4 \\
Expansion factor & 2 \\
Kernel size & 4 \\
Use bias in convolutional layers & True \\
Dropout rate & 0.0 \\
Dimensionality of the state & 128 \\
Normalization method & RMSNorm \\
Activation function & SiLU \\
PLE encoding dimension & 64 \\
Axis & 1 \\
\bottomrule
\end{tabular}
\label{tab:mambatab_hyperparameters}
\end{table}

Additionally, we test out the default architecture from \citet{ahamed2024mambatab} and hence have a super small model with only a single layer and embedding dimensionality of 32. 

\begin{table}[ht]
\centering
\caption{Default Hyperparameters for the MambaTab Model}
\begin{tabular}{ll}
\toprule
\textbf{Hyperparameter} & \textbf{Value} \\
\midrule
Model Dim & 32 \\
Number of layers & 1 \\
Expansion factor & 2 \\
Kernel size & 4 \\
Use bias in convolutional layers & True \\
Dropout rate & 0.0 \\
Dimensionality of the state & 32 \\
Normalization method & RMSNorm \\
Activation function & SiLU \\
Axis & 1 \\
\bottomrule
\end{tabular}
\label{tab:mambatab_hyperparameters}
\end{table}

Lastly, we follow the Github implementation from \citet{ahamed2024mambatab} where the sequence is flipped and the SSM iterates over the number of observations instead of the pseudo-sequence length of 1.
\begin{table}[ht]
\centering
\caption{Default Hyperparameters for the MambaTab$^T$ Model}
\begin{tabular}{ll}
\toprule
\textbf{Hyperparameter} & \textbf{Value} \\
\midrule
Model Dim & 32 \\
Number of layers & 1 \\
Expansion factor & 2 \\
Kernel size & 4 \\
Use bias in convolutional layers & True \\
Dropout rate & 0.0 \\
Dimensionality of the state & 32 \\
Normalization method & RMSNorm \\
Activation function & SiLU \\
Axis & 0 \\
\bottomrule
\end{tabular}
\label{tab:mambatab_hyperparameters}
\end{table}

\paragraph{Mambular} For Mambular we create a sensible default, following hyperparameters from the literature. We keep all hyperparameters from the Mambablocks as introduced by \citet{gu2023mamba}. Hence we use SiLU activation and RMSNorm. WE use an expansion factor of 2 and use an embedding dimensionality of 64. The PLE encoding dimension is adapted to always match the embedding dimensionalitiy since first expanding the dimensionality in preprocessing to subsequently reduce it in the embedding layer seems counter intuitive.

\begin{table}[ht]
\centering
\caption{Default Hyperparameters for the Mambular Model}
\begin{tabular}{ll}
\toprule
\textbf{Hyperparameter} & \textbf{Value} \\
\midrule
Model Dim & 64 \\
Number of layers & 4 \\
Expansion factor & 2 \\
Kernel size & 4 \\
Use bias in convolutional layers & True \\
Dropout rate & 0.0 \\
Dimensionality of the state & 128 \\
Normalization method & RMSNorm \\
Activation function & SiLU \\
PLE encoding dimension & 64 \\
\bottomrule
\end{tabular}
\label{tab:mambatab_hyperparameters}
\end{table}

\paragraph{Model sizes} Below you find the number of trainable parameters for all models for all datasets. Note, that MambaTab$^*$ and Mambular have very similar numbers of parameters since the sequence length does not have a large impact on the number of model parameters. Overall there is no correlation between model size and performance since e.g. the FT-Transformer architecture which is comparably larger to e.g. the MLP and ResNet architectures performs very well whereas the largest architecture, the TabTransformer performs worse than the smaller ResNet. Additionally, since the models have distinctively different architectures, the overall number of trainable parameters is not conclusive for training time or memory usage.

\begin{table}[ht]
    \centering
    \caption{Number of trainable parameters for all models and datasets. Note that the number of trainable parameters is dependent on the dataset, since e.g. a larger number of variables leads to more trainable parameters in the embedding layer.}
    \resizebox{\textwidth}{!}{%
\begin{tabular}{l|cccccccccccc}
\toprule
Dataset        &     AB &     AD &     BA &     CA &     CH &    CP &     DI &     FI &     HS &     MA &     PA &    WI \\
\midrule
FT-Transformer &   765k &   709k &   795k &   784k &   722k &  852k &   763k &   834k &   837k &   794k &   944k &  822k \\
MLP            &   242k &   103k &   124k &   280k &   156k &  418k &   233k &   351k &   310k &   105k &   594k &  356k \\
ResNet         &   261k &   123k &   144k &   299k &   176k &  437k &   253k &   371k &   330k &   125k &   614k &  375k \\
TabTransformer &  1060k &  1073k &  1149k &  1061k &  1060k &   -   &  1063k &     -  &  1100k &  1157k &  1068k &   - \\
MambaTab$^*$   &   331k &   318k &   316k &   335k &   321k &  352k &   328k &   358k &   339k &   312k &   373k &  348k \\
MambaTab       &    13k &    14k &    14k &    13k &    13k &   14k &    13k &    14k &    14k &    14k &    14k &   14k \\ 
Mambular       &   331k &   324k &   361k &   335k &   321k &  352k &   329k &   365k &   358k &   361k &   374k &  348k \\
\bottomrule
\end{tabular}
}
\end{table}

\section{Additional Results}\label{app:add_results}

Additional results, including a comparison to MambaTab and the presented variants from the supplemental material above are given below.

\begin{table}[htbp]
\centering
\caption{Benchmarking results for the regression tasks.  Average mean squared error values over 5 folds and the corresponding standard deviations are reported. Smaller values are better. The best performing model is marked in bold. 
}
\resizebox{\textwidth}{!}{%
\begin{tabular}{l|ccccccc}
\toprule
Models  & DI $\downarrow$ & AB $\downarrow$ & CA $\downarrow$ & WI $\downarrow$ & PA $\downarrow$ & HS $\downarrow$ & CP $\downarrow$ \\
\midrule
XGBoost         & 0.019 \std{0.000} & 0.506 \std{0.044} & 0.171 \std{0.007} & 0.528 \std{0.008} & 0.036 \std{0.004} & 0.119 \std{0.024} & 0.024 \std{0.004} \\
RF              & 0.019 \std{0.001} & 0.461 \std{0.052} & 0.183 \std{0.008} & \textbf{0.485} \std{0.007} & 0.028 \std{0.006} & 0.121 \std{0.018} & 0.025 \std{0.002} \\
LightGBM        & 0.019 \std{0.001} & 0.459 \std{0.047} & 0.171 \std{0.007} & 0.542 \std{0.013} & 0.039 \std{0.007} & 0.112 \std{0.018} & 0.023 \std{0.003} \\
CatBoost       &  0.019 \std{0.000}  & 0.457 \std{0.007} & 0.169 \std{0.006} & 0.583 \std{0.006} & 0.045 \std{0.006} & \textbf{0.106} \std{0.015} & \textbf{0.022} \std{0.001} \\
FT-Transformer  & 0.018 \std{0.001} & 0.458 \std{0.055} & 0.169 \std{0.006} & 0.615 \std{0.012} & \textbf{0.024} \std{0.005} & 0.111 \std{0.014} & 0.024 \std{0.001} \\
MLP             & 0.066 \std{0.003} & 0.462 \std{0.051} & 0.198 \std{0.011} & 0.654 \std{0.013} & 0.764 \std{0.023} & 0.147 \std{0.017} & 0.031 \std{0.001} \\
TabTransformer  & 0.065 \std{0.002} & 0.472 \std{0.057} & 0.247 \std{0.013} & -                 & 0.135 \std{0.001} & 0.160 \std{0.028} & - \\
ResNet          & 0.039 \std{0.018} & 0.455 \std{0.045} & 0.178 \std{0.006} & 0.639 \std{0.013} & 0.606 \std{0.031} & 0.141 \std{0.017} & 0.030 \std{0.002} \\
NODE            & 0.019 \std{0.000} &  \textbf{0.431} \std{0.052} & 0.207 \std{0.001} & 0.613  \std{0.006} & 0.045 \std{0.007}   & 0.124  \std{0.015} & 0.026 \std{0.001} \\  
LinReg          & 0.115 \std{0.002} & 0.483 \std{0.055} & 0.365 \std{0.021} & 0.711 \std{0.006} & 0.830 \std{0.047} & 0.302 \std{0.033} & 0.289 \std{0.004} \\
MambaTab        & 0.035 \std{0.006} & 0.456 \std{0.053} & 0.272 \std{0.016} & 0.685 \std{0.015} & 0.531 \std{0.032} & 0.163 \std{0.009} & 0.030 \std{0.002} \\
MambaTab$^{T}$  & 0.038 \std{0.002} & 0.468 \std{0.048} & 0.279 \std{0.010} & 0.694 \std{0.015} & 0.576 \std{0.022} & 0.179 \std{0.027} & 0.033 \std{0.002} \\
MambaTab$^*$    & 0.040 \std{0.008} & 0.455 \std{0.043} & 0.180 \std{0.008} & 0.601 \std{0.010} & 0.571 \std{0.021} & 0.122 \std{0.017} & 0.030 \std{0.002} \\
\hline
Mambular        & \textbf{0.018} \std{0.000} & 0.452 \std{0.043} & \textbf{0.167} \std{0.011} & 0.628 \std{0.010} & 0.035 \std{0.005} & 0.132 \std{0.020} & 0.025 \std{0.002} \\
\bottomrule
\end{tabular}
}

\label{tab:reg_regression_all}
\end{table}

\begin{table}[htbp]
\centering
\caption{Benchmarking results for the classification tasks.  Average AUC values over 5 folds and the corresponding standard deviations are reported. Larger values are better.}
\resizebox{\textwidth}{!}{%
\begin{tabular}{l|ccccc}
\toprule
Models          & BA $\uparrow$ & AD $\uparrow$ & CH $\uparrow$ & FI $\uparrow$ & MA $\uparrow$\\
\midrule

XGBoost         & 0.928 \std{0.004} & \textbf{0.929} \std{0.002} & 0.845 \std{0.008} & 0.774 \std{0.009} & 0.922 \std{0.002} \\
RF              & 0.923 \std{0.006} & 0.896 \std{0.002} & 0.851 \std{0.008} & 0.789 \std{0.012} & 0.917 \std{0.004} \\
LightGBM        & \textbf{0.932} \std{0.004} & 0.929 \std{0.001} & 0.861 \std{0.008} & 0.788 \std{0.010} & \textbf{0.927} \std{0.001} \\
CatBoost        & 0.932 \std{0.008} & 0.927 \std{0.002}  & \textbf{0.867} \std{0.006} &0.796 \std{0.010} &   0.926  \std{0.005}\\
FT-Transformer  & 0.926 \std{0.003} & 0.926 \std{0.002} & 0.863 \std{0.007} & 0.792 \std{0.011} & 0.916 \std{0.003} \\
MLP             & 0.895 \std{0.004} & 0.914 \std{0.002} & 0.840 \std{0.005} & 0.793 \std{0.011} & 0.886 \std{0.003} \\
TabTransformer  & 0.921 \std{0.004} & 0.912 \std{0.002} & 0.835 \std{0.007} & - & 0.910 \std{0.002} \\
ResNet          & 0.896 \std{0.006} & 0.917 \std{0.002} & 0.841 \std{0.006} & 0.793 \std{0.013} & 0.889 \std{0.003} \\
NODE            & 0.914 \std{0.008} & 0.904 \std{0.002} & 0.851 \std{0.006} & 0.790 \std{0.010}  & 0.904 \std{0.005}  \\
Log-Reg         & 0.810 \std{0.008} & 0.838 \std{0.001} & 0.754 \std{0.006} & 0.768 \std{0.013} & 0.800 \std{0.005} \\
MambaTab$^*$    & 0.900 \std{0.004} & 0.916 \std{0.003} & 0.846 \std{0.007} & 0.792 \std{0.011} & 0.890 \std{0.003} \\
MambaTab        & 0.886 \std{0.006} & 0.901 \std{0.001} & 0.828 \std{0.005} & 0.785 \std{0.012} & 0.880 \std{0.003} \\
MambaTab$^{T}$  & 0.888 \std{0.005} & 0.899 \std{0.002} & 0.815 \std{0.009} & 0.783 \std{0.012} & 0.878 \std{0.005} \\
\hline
Mambular        & 0.927 \std{0.006} & 0.928 \std{0.002} & 0.861 \std{0.008} & \textbf{0.796} \std{0.013} & 0.917 \std{0.003} \\
\bottomrule
\end{tabular}
}

\label{tab:classification-results-all}
\end{table}

\begin{table}[ht]
\centering
\caption{Combined Ranking of Models for Regression and Classification Tasks}
\begin{tabular}{l|cc||c}
\toprule
Models & Regression Rank & Classification Rank & Overall Rank \\
\midrule

XGBoost             &    5.14   \std{4.02} &  5.4  \std{4.51}   &  5.25   \std{4.03} \\
RF                  &    5.00   \std{2.94} &  7.4  \std{3.36}   &  6.00   \std{3.22} \\
LightGBM            &    4.57   \std{2.07} &  3.4  \std{3.36}   &  4.08   \std{2.61} \\
CatBoost            &    4.00   \std{2.45} &  \textbf{2.2}  \std{1.10}   &  \textbf{3.25}   \std{2.14} \\
FT-Transformer      &    \textbf{3.57}   \std{2.51} &  4.6  \std{1.52}   &  4.00   \std{2.13} \\
MLP                 &   10.86   \std{1.57} &  8.6  \std{3.36}   &  9.92   \std{2.61} \\
TabTransformer      &   10.80   \std{1.64} &  8.5  \std{1.91}   &  9.78   \std{2.05} \\
ResNet              &    8.14   \std{2.91} &  7.6  \std{3.05}   &  7.92   \std{2.84} \\
\hline
NODE                &    5.71   \std{2.93} &  7.6  \std{1.82}   &  6.50   \std{2.61} \\
Regression          &   13.57   \std{0.53} & 13.8  \std{0.45}   &  13.67  \std{0.49} \\
MambaTab            &    9.29   \std{2.56} & 11.6  \std{1.14}   &  10.25  \std{2.34} \\
MambaTab$^{T}$      &   11.57   \std{1.40} & 12.2  \std{0.84}   &  11.83  \std{1.19} \\
MambaTab$^*$        &    7.11   \std{2.79} &  7.4  \std{1.67}   &  7.25   \std{2.30} \\     
Mambular            &    4.00   \std{3.06} &  3.0  \std{1.22}   &  3.58   \std{2.43} \\
\bottomrule
\end{tabular}
\label{tab:averages_with_std}
\end{table}

\end{document}